\documentclass{article}

\usepackage[preprint]{neurips_2024}

\usepackage[utf8]{inputenc} 
\usepackage[T1]{fontenc}    
\usepackage{hyperref}       
\usepackage{url}            
\usepackage{booktabs}       
\usepackage{amsfonts}       
\usepackage{nicefrac}       
\usepackage{microtype}      
\usepackage{graphicx}
\usepackage{wrapfig, lipsum}
\usepackage[dvipsnames]{xcolor}
\usepackage{pifont}
\usepackage{enumitem}
\usepackage{color}

\newcommand{\cmark}{\checkmark}
\newcommand{\xmark}{\scalebox{0.75}{\usym{2613}}}

\usepackage[many]{tcolorbox}
\usepackage{multirow}
\usepackage{utfsym}
\usepackage{tablefootnote}

\title{Chat-Scene: Bridging 3D Scene and Large Language Models with Object Identifiers}

\author{
\small
    Haifeng Huang$^{1,2}$$^*$ \hspace{0.8em} Yilun Chen$^{2}$$^*$ \hspace{0.8em} Zehan Wang$^{1}$\thanks{Equal contribution.} \hspace{0.8em} Rongjie Huang$^{1}$ \hspace{0.8em} Runsen Xu$^{2}$ \hspace{0.8em} Tai Wang$^{2}$ \\
\small
    \textbf{Luping Liu}$^{1}$ \hspace{1em} \textbf{Xize Cheng}$^{1}$ \hspace{1em} \textbf{Yang Zhao}$^{3}$ 
    \hspace{1em} \textbf{Jiangmiao Pang}$^{2}$ \hspace{1em} \textbf{Zhou Zhao}$^{1,2}$\thanks{Corresponding author.}\\
\\
\small
    $^{1}$Zhejiang University \hspace{1em} $^{2}$Shanghai AI Laboratory \hspace{1em} $^{3}$Bytedance Inc.\\
    {\tt\small \{huanghaifeng\}@zju.edu.cn}
}

\begin{document}

\maketitle

\begin{abstract}
Recent advancements in 3D Large Language Models (LLMs) have demonstrated promising capabilities for 3D scene understanding. However, previous methods exhibit deficiencies in general referencing and grounding capabilities for intricate scene comprehension. In this paper, we introduce the use of object identifiers and object-centric representations to interact with scenes at the object level. Specifically, we decompose the input 3D scene into a set of object proposals, each assigned a unique identifier token, which enables efficient object referencing and grounding during user-assistant interactions. Given the scarcity of scene-language data, we model the scene embeddings as a sequence of explicit object-level embeddings, derived from semantic-rich 2D or 3D representations. By employing object identifiers, we transform diverse 3D scene-language tasks into a unified question-answering format, facilitating joint training without the need for additional task-specific heads. With minimal fine-tuning on all downstream tasks, our model significantly outperforms existing methods on benchmarks including ScanRefer, Multi3DRefer, Scan2Cap, ScanQA, and SQA3D.
\end{abstract}

\section{Introduction}
\label{sec:intro}
Recent advancements in Large Language Models (LLMs) \cite{chiang2023vicuna, gpt4, touvron2023llama, chowdhery2022palm, ye2023mplug, li2023otter} have established language as a universal interface for creating general-purpose assistants. This breakthrough has been instrumental in the development of Multi-modal LLMs (MLLMs), which effectively tackle a broad spectrum of multi-modal tasks. While significant strides have been made in 2D MLLMs \cite{li2023videochat, llava, zhao2023bubogpt, zhu2023minigpt, llava1-5, lisa, ferret}, current 3D MLLMs still face significant challenges that must be overcome to achieve a general-purpose assistant for 3D scene understanding.

Object referencing and grounding are essential for advanced scene understanding. Object referencing involves a model's precise comprehension of the semantics associated with a user-specified object, while object grounding requires the model's ability to localize a target object within the scene. These capabilities are vital for various 3D scene-language tasks such as dense captioning~\cite{scan2cap} and visual grounding~\cite{scanrefer, multi3drefer, referit3d}. However, current 3D MLLMs lack general referencing and grounding capabilities, often failing in tasks that necessitate precise object referencing or grounding—contrary to the objectives of addressing general-purpose tasks.

Regarding the object referencing capability, several 3D MLLMs~\cite{ll3da, leo, chat3d} employ additional prompt encoders to comprehend user-specified objects, but they still lack grounding capabilities. The 3D-LLM~\cite{3dllm} incorporates location tokens to enable object grounding, a technique validated in the 2D domain~\cite{ferret}. However, this approach underperforms on the 3D grounding benchmark, ScanRefer~\cite{scanrefer}, compared to traditional expert models. The ineffectiveness of location tokens in the 3D domain primarily arises from the significant data scarcity in the scene-language area. Current 3D scene-language datasets~\cite{scanrefer, multi3drefer, referit3d} contain only tens of thousands of grounding instances, a scale much smaller than the million-level datasets used for training 2D MLLMs~\cite{ferret, ferretv2}. Given the exponentially greater complexity of 3D spaces compared to 2D spaces, robust training of location tokens for 3D MLLMs may require substantially more data than is currently used for 2D MLLMs. Therefore, our objective is to explore more efficient methods for object referencing and grounding and to mitigate the impact of data scarcity.

We observe that most existing 3D MLLMs convert the 3D scene into hidden 3D scene embeddings, employing either a Q-Former-based module~\cite{3dllm, ll3da} or direct projection methods~\cite{leo}. Such architectures inherently lack the capability to efficiently interpret individual object instances. To address this limitation, we propose a novel approach for \textbf{representing and interacting with 3D scenes at the object level} within the language model. This method incorporates two principal designs: (i) referencing 3D scene using object identifiers, and (ii) representing 3D scene using well-trained object-centric representations. The first component offers a unified format for object referencing and grounding, while the second alleviates the requirement for extensive scene-language datasets.

\setlength{\intextsep}{3pt}
\begin{wrapfigure}{r}{0.5\textwidth}
    \centering
    \includegraphics[width=0.49\textwidth]{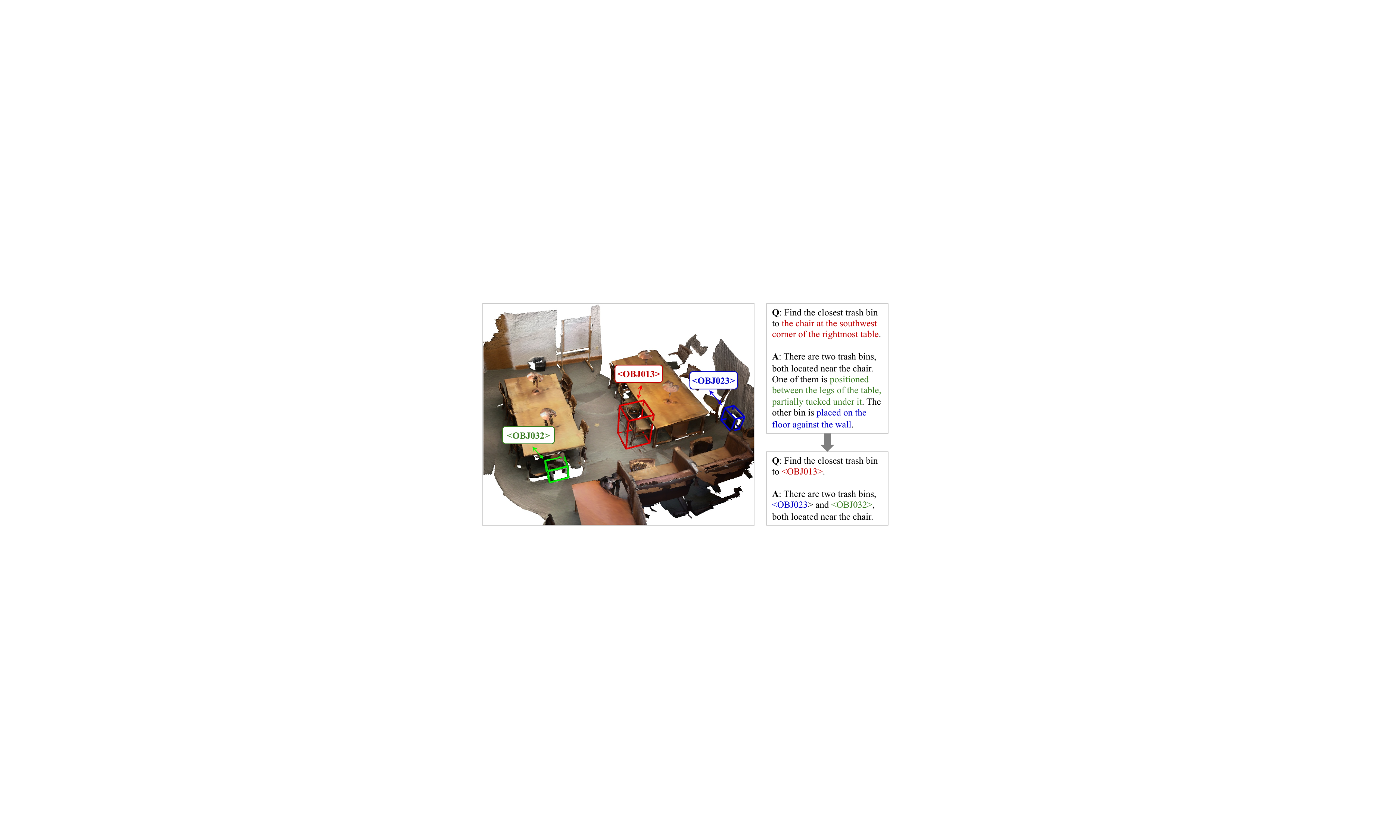}
    \caption{An example of using object identifiers during the conversation.}
    \label{fig:intro}
\end{wrapfigure}

\noindent\textbf{Reference 3D scene using object identifiers.} Objects play a crucial role in defining and interpreting a scene, as their organization shapes the entire 3D landscape. This intuition is evident in most 3D scene understanding benchmarks, including 3D grounding, VQA, and dense captioning, all of which annotate at the object level. To effectively model scene embeddings at the object level, the entire 3D scene can be decomposed into a set of object proposals via reliable 3D detectors~\cite{mask3d, pointgroup, sam3d, softgroup, oneformer3d}. Importantly, we assign objects with object identifiers—a set of learnable identifier tokens $\left \{\texttt{<OBJ}k\texttt{>} \right\}_{k=1...n}$—to distinguish them during language modeling. This design allows the LLM to reference respective objects using discrete identifier tokens. As the example shown in Figure~\ref{fig:intro}, the chair and the two trash cans are labeled as ``$\texttt{<OBJ013>}$'', ``$\texttt{<OBJ023>}$'', and ``$\texttt{<OBJ032>}$'', respectively. This avoids the text ambiguity that arises from subjective viewing words like ``rightmost''. Besides, the lengthy description like ``the chair located at the southwest corner of the rightmost table'' often complicates user-assistant interaction. These identifiers enable efficient object referencing and grounding during user-assistant interactions. By using these identifiers, we convert diverse 3D scene-language tasks into a unified format of question-answering pairs, facilitating joint training without any additional task-specific heads.

\noindent\textbf{Represent 3D scene using well-trained object-centric representations.}
The scene-level representation requires a large amount of paired scene-language data for training, which is generally unaffordable and labor-intensive due to its complex real-world scenarios. To address this challenge, our model represents the scenes using a set of object-level embeddings, which obtain object-centric representations from well-trained 2D and 3D models. Specifically, after obtaining object proposals from prior detectors (either 2D or 3D), we extract the object features using well-trained 3D object-centric representations~\cite{uni3d} or 2D representations~\cite{dinov2}. Due to the million-level pre-training, these representations contain abundant semantic and visual cues. Through simple linear layers, we project them into the embedding space of the language model. Combined with the object identifier, the sequences of object-level embeddings are thus constructed into scene embeddings and fed into the LLM. With merely \textit{two} epochs of fine-tuning on all downstream tasks, extensive experiments on either 3D, 3D+2D, or 2D-only settings demonstrate the effectiveness of our model in various downstream 3D scene understanding tasks.

We perform comprehensive experiments across five representative 3D scene-language datasets, including ScanRefer~\cite{scanrefer}, Multi3DRefer~\cite{multi3drefer}, Scan2Cap~\cite{scan2cap}, ScanQA~\cite{scanqa}, and SQA3D~\cite{sqa3d}. Our model consistantly enhances state-of-the-art performance across all these datasets without fine-tuning on specific tasks. Notably, it surpasses previous methods by 3.7\% (Acc@0.5) on ScanRefer, 14.0\% (F1@0.5) on Multi3DRefer, 8.7\% (CIDEr@0.5) on Scan2Cap, and 8.4\% (CIDEr) on ScanQA.


Our contributions are summarized as follows:
\begin{itemize}[leftmargin=*]
\setlength{\itemsep}{0pt}
\setlength{\parskip}{0pt}
\item We propose an enhanced 3D MLLM which models and interacts with 3D scenes at the object level.
\item We introduce object identifiers to enable efficient referencing and grounding within 3D scenes. By leveraging these identifiers, we convert diverse 3D scene-language tasks into a unified question-answering format, facilitating joint training without necessitating additional task-specific heads.
\item We effectively represent the 3D scene through a sequence of multi-modal object-centric representations derived from well-trained foundation models, which alleviate the impact of scene-language data scarcity.
\item Our model significantly enhances state-of-the-art performance across various 3D scene-language datasets without fine-tuning for specific tasks. Extensive experiments on either 3D, 3D+2D, or 2D-only settings demonstrate the effectiveness of our model for 3D indoor scene understanding.
\end{itemize}

\section{Related Work}
\label{sec:formatting}

\noindent\textbf{3D Scene-language Understanding}
In the rapidly evolving field of 3D scene understanding, there is an increasing focus on using language to provide both contextual knowledge and query conditions, thus enabling precise interpretation of user intentions. This process, known as ``3D scene-language understanding'', leverages language to more effectively grasp the intricacies of 3D environments in a manner consistent with human cognition. The primary tasks in this domain include:  1) 3D Visual Grounding~\cite{scanrefer, referit3d, mvt, multi3drefer, vil3drel, 3drp-net, 3dvg-transformer, wang2023distilling, concretenet}, which involves locating specified objects within a 3D scene based on textual queries; 2) 3D Dense Captioning~\cite{scan2cap, X-trans2cap, more, vote2cap, vote2cap++}, which demands proficiency in both localizing and captioning objects densely in the scene; 3) 3D Visual Question Answering~\cite{scanqa, parelli2023clip, sqa3d}, which focuses on general scene question answering. Initial efforts concentrated on specialized tasks, resulting in limited generalizability across different 3D scene understanding tasks. Recent initiatives such as 3DJCG~\cite{3djcg} and D3Net~\cite{d3net} have aimed to unify tasks like 3D visual grounding and dense captioning, leveraging their synergistic benefits to enhance overall model performance. Advances like 3D-VisTA~\cite{3dvista} and 3D-VLP~\cite{3d-vlp} are working to develop a more general 3D visual-language framework through pre-training techniques for better scene-language alignment. However, despite these models' adeptness at handling various 3D scene tasks, their reliance on task-specific heads limits their adaptability for broader user-assistant interactions.

\noindent\textbf{Multi-modal Large Language Models.} 
Recent advancements in large language models (LLMs) have exhibited impressive capabilities in intricate reasoning and interactive dialogues with humans~\cite{chiang2023vicuna, gpt4, touvron2023llama, chowdhery2022palm}. There is a growing interest in enhancing the scope of LLMs to encompass additional modalities~\cite{lisa, li2023videochat, llava, zhao2023bubogpt, zhu2023minigpt, han2023imagebind, pointbind, 3dllm, chat3d, ye2023mplug, li2023otter, pointllm, leo}. In the 3D realm, PointLLM~\cite{pointllm} directly maps point clouds into the embedding space of the LLM. Both Imagebind-LLM~\cite{han2023imagebind} and Point-LLM~\cite{pointbind} integrate the 3D modality into LLMs by establishing a joint embedding space among 3D point clouds, images, audio, and text. These models perform well at the object level but encounter difficulties when interpreting complex spatial relationships in 3D scenes. To improve scene understanding, 3D-LLM~\cite{3dllm} incorporates positional embedding and learns location tokens. Nevertheless, it projects 3D features into the input space of pre-trained 2D Vision-Language Models (VLMs). Involving 2D encoders make it difficult to grasp the 3D spatial structure and intricate relationships among objects. Chat-3D~\cite{chat3d} tackles this limitation by directly utilizing 3D scene-text data to align the 3D scene with the LLM, overcoming the challenge of limited data availability through a pre-alignment phase. However, the architectural design of this model limits its focus on specific target objects during interactions. Current 3D MLLMs face challenges in precise object referencing and grounding, limiting their functionality to straightforward tasks. By incorporating object identifiers into the LLM, we significantly enhance the object referencing and grounding capabilities of 3D MLLMs, thereby showing potential for complex real-world applications.

\section{Method\label{sec:method}}

\begin{figure}[t]
    \centering
    \includegraphics[width=\textwidth]{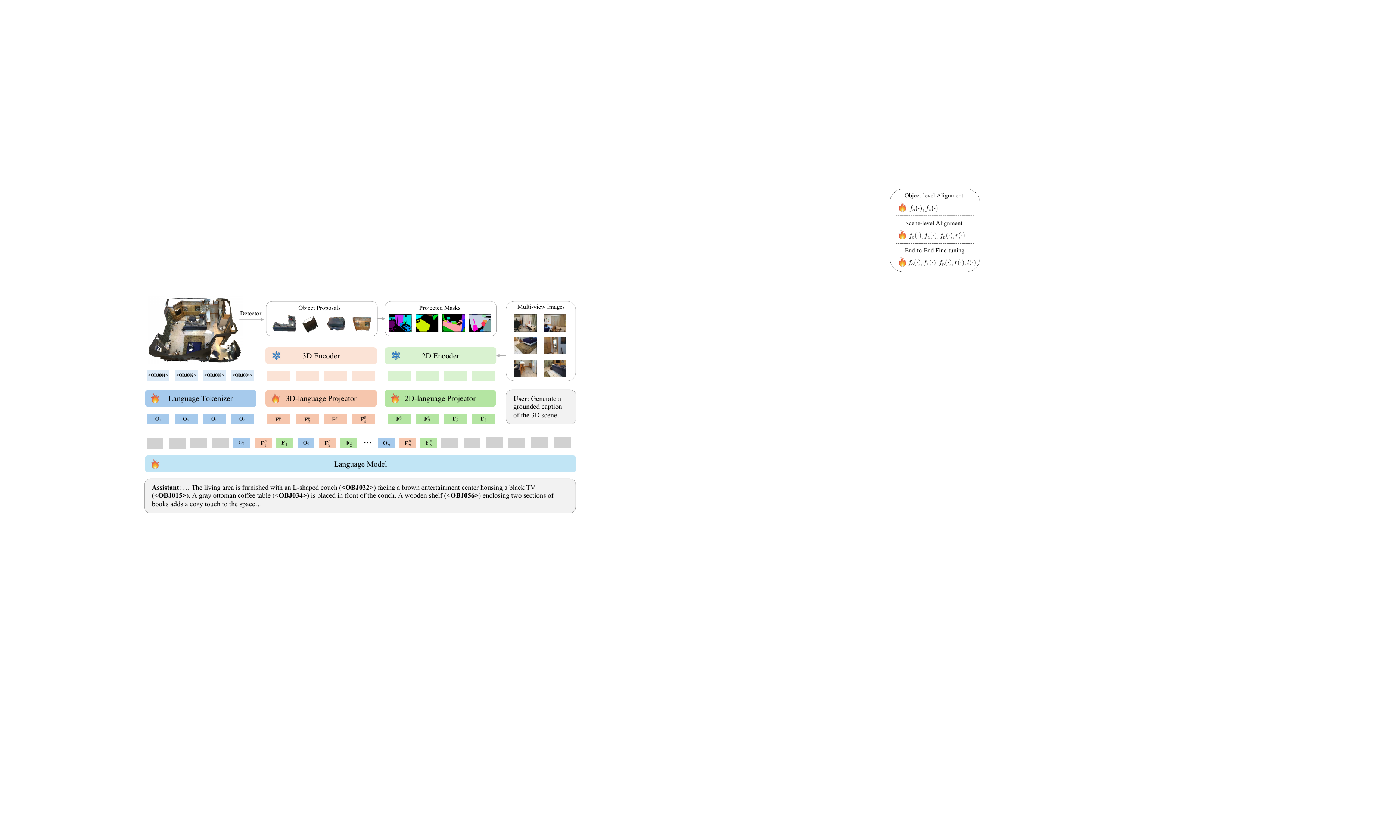}
    \caption{\textbf{Overall model architecture.} The model processes a 3D scene's point cloud input by decomposing it into object proposals via a pre-trained detector. Subsequently, the 3D and 2D encoders are employed to extract object-centric representations. After projection layers, they are combined with object identifiers to form the scene embeddings as a sequence of object-level embeddings, which are then fed into the LLM. The assigned unique identifiers enable efficient object referencing in subsequent interactions.}
    \label{fig:model}
\end{figure}

\subsection{Overview}

Our motivation is to facilitate object referencing and grounding for 3D MLLMs while simultaneously addressing the scarcity of scene-language data. We propose representing 3D scenes at the object level by using object identifiers for referencing and employing well-trained, object-centric representations for scene depiction. Section~\ref{sec:model_arch} delineates the model architecture, which primarily consists of generating a sequence of object-level embeddings to represent the entire scene. Section~\ref{sec:prompt_template} provides illustrations of the prompt template through examples. Lastly, Section~\ref{sec:training_strategy} details the training methodology of our model.

\subsection{Model Architecture\label{sec:model_arch}}
As illustrated in Figure~\ref{fig:model}, our method processes a 3D scene's point cloud by decomposing it into object proposals using a pre-trained detector. We then employ pre-trained 3D and 2D encoders to derive object-centric representations from point clouds and multi-view images, respectively. These representations are subsequently mapped into the token embedding space of a language model. By incorporating $n$ object identifiers into the language model's vocabulary, we link these identifiers to the corresponding object proposals, thereby facilitating efficient object referencing and grounding during user-assistant interactions. Finally, the scene embeddings, composed of a sequence of object-level embeddings, are input into the LLM.

\noindent\textbf{Object Detector.}
Given a point cloud from a 3D scene, we decompose it into $n$ objects using the pre-trained detector Mask3D~\cite{mask3d}. Compared to object detection models~\cite{focalformer3d, v-detr}, the instance segmentation model is preferred in this work due to due to its capability to generate accurate masks, which are essential for subsequent projection into 2D masks on multi-view images. The point cloud of the $i$-th object is denoted as $\mathbf{P}_i\in \mathbb{R}^{m_i\times 6}$, where $m_i$ represents the number of points in the $i$-th object, and the 6D information for each point comprises its 3D coordinates and RGB colors.

\noindent\textbf{Object Identifiers.}
To achieve a localized understanding of 3D scenes, we introduce a set of learnable identifier tokens $\left \{\texttt{<OBJ}i\texttt{>} \right\}_{i=1...n}$, designated as object identifiers, into the token vocabulary of the language model. These identifiers are processed by the tokenizer to produce their respective token embeddings $\left \{\mathbf{O}_i \right\}_{i=1...n}$. The identifier tokens are then integrated with the object tokens to establish one-to-one correspondences, thereby enabling object referencing and grounding using identifiers in subsequent interactions.

\noindent\textbf{Object-level Embeddings.}
After extracting object proposals from the 3D scene, we derive object features using well-trained 3D and 2D object representations. Owing to the million-level scale of pre-training, these representations are rich in semantic and visual cues. By employing simple linear layers, we project them into the embedding space of the language model. Together with identifier token embeddings, this process yields object-level embeddings for each object.

\noindent\textit{\textbf{3D Encoder.}}
The 3D encoder excels in extracting spatial and shape attributes from point clouds. We employ a pre-trained 3D encoder Uni3D~\cite{uni3d} to derive object-centric 3D representations. This embedding processes each object's point cloud $\mathbf{P}_i$, outputting the feature $\mathbf{Z}^{p}_{i}$ for each object.

\noindent\textit{\textbf{2D Encoder.}}
The 2D encoder adeptly extracts semantically rich features from 2D images. We project the point clouds for each object onto multi-view images, creating a sequence of 2D masks. Utilizing a pre-trained DINOv2~\cite{dinov2}, we extract and aggregate local features from all masked regions across the multi-view images of each object, taking into account both mask areas and multi-view information. We opted for DINOv2 over the more common CLIP~\cite{parelli2023clip} due to its superior handling of local features within images. The 2D encoder processes the multi-view images and their corresponding projected masks from each object's point cloud $\mathbf{P}_i$, generating the visual feature $\mathbf{Z}^v_i$ for each object. Details are provided in Appendix~\ref{appendix:implementation}.

\noindent\textit{\textbf{Visual-Language Projectors.}}
To align the extracted object representations with the language model, we employ a 3D-language projector $f_p(\cdot)$ and a 2D-language projector $f_v(\cdot)$ to map the 3D point cloud features and 2D visual features into the token embedding space of the language model. For the $i$-th object, these features are represented as token embeddings $\mathbf{F}^{p}_{i}$ and $\mathbf{F}^{v}_{i}$.
\begin{equation}
\mathbf{F}^{p}_{i} = f_p(\mathbf{Z}^{p}_{i});\quad \mathbf{F}^{v}_{i} = f_v(\mathbf{Z}^{v}_{i}).
\end{equation}
\noindent\textbf{Scene Embeddings.}
Following the process described above, we obtain an object identifier token embedding $\mathbf{O}_i$, a 3D object token embedding $\mathbf{F}^{p}_{i}$, and a 2D object token embedding $\mathbf{F}^{v}_{i}$ for each object. We combine the identifier token embeddings and object token embeddings in an one-to-one correspondence manner to formulate a sequence of object-level embeddings, which represents the constructed scene embeddings and then be fed into the LLM to represent the whole scene.

\subsection{Prompt Template\label{sec:prompt_template}}
Despite variations in task formulations, both referencing and grounding are unified using object identifiers. As illustrated in Table~\ref{tab:prompt_template}, the system message encodes object information in the scene as a sequence of ``$\texttt{<OBJ}i\texttt{>}$ $\texttt{<object>}$'', where $\texttt{<OBJ}i\texttt{>}$ denotes the identifier token for the $i$-th proposal, and $\texttt{<object>}$ serves as the placeholder for object tokens. The language tokenizer converts $\texttt{<OBJ}i\texttt{>}$ into its token embedding $\mathbf{O}_i$ and $\texttt{<object>}$ into the combined object token features $\mathbf{F}^{p}_{i}$ and $\mathbf{F}^{v}_{i}$. As illustrated by the following interaction, the user can directly employ identifier tokens to reference specific objects, while the assistant uses these tokens in responses to precisely ground target objects.

\begin{table*}[htbp]
\caption{\textbf{Prompt template for the language model.} }
\label{tab:prompt_template}
\begin{tcolorbox}[colback=lightgray!10,
                    colframe=black,
                    width=\textwidth,
                    arc=1mm, auto outer arc,
                    boxrule=0.5pt,
                    ]
\textbf{System}: A chat between a curious user and an artificial intelligence assistant. The assistant gives helpful, detailed, and polite answers to the user's questions. The conversation centers around an indoor scene: [$\texttt{<OBJ001>}$ $\texttt{<object>}$ $\texttt{<OBJ002>}$ $\texttt{<object>}$ ... $\texttt{<OBJ}n\texttt{>}$ $\texttt{<object>}$]. 

\textbf{User:} Find the closest trash bin to $\texttt{<OBJ013>}$.

\textbf{Assistant:} There are two trash bins, $\texttt{<OBJ023>}$ and $\texttt{<OBJ032>}$, both located near the chair.

\end{tcolorbox}
\label{tab:prompt_input}
\end{table*}

\subsection{Training Strategy\label{sec:training_strategy}}
Most existing MLLMs~\cite{scenellm, llava, leo} adopt a two-stage training approach, comprising an initial alignment phase to train the projector exclusively, followed by a fine-tuning phase for both the projector and the language model. This method not only demands extra data and extended training duration for alignment but also complicates determining the optimal duration for the initial phase. Consequently, we opt for a single-stage process, concurrently training both the projectors and the language model. In our experiments, we observe that this jointly trained model already exhibits superior performance without the necessity of fine-tuning for specific downstream tasks.

\begin{figure*}[tbp]
    \centering
    \includegraphics[width=\textwidth]{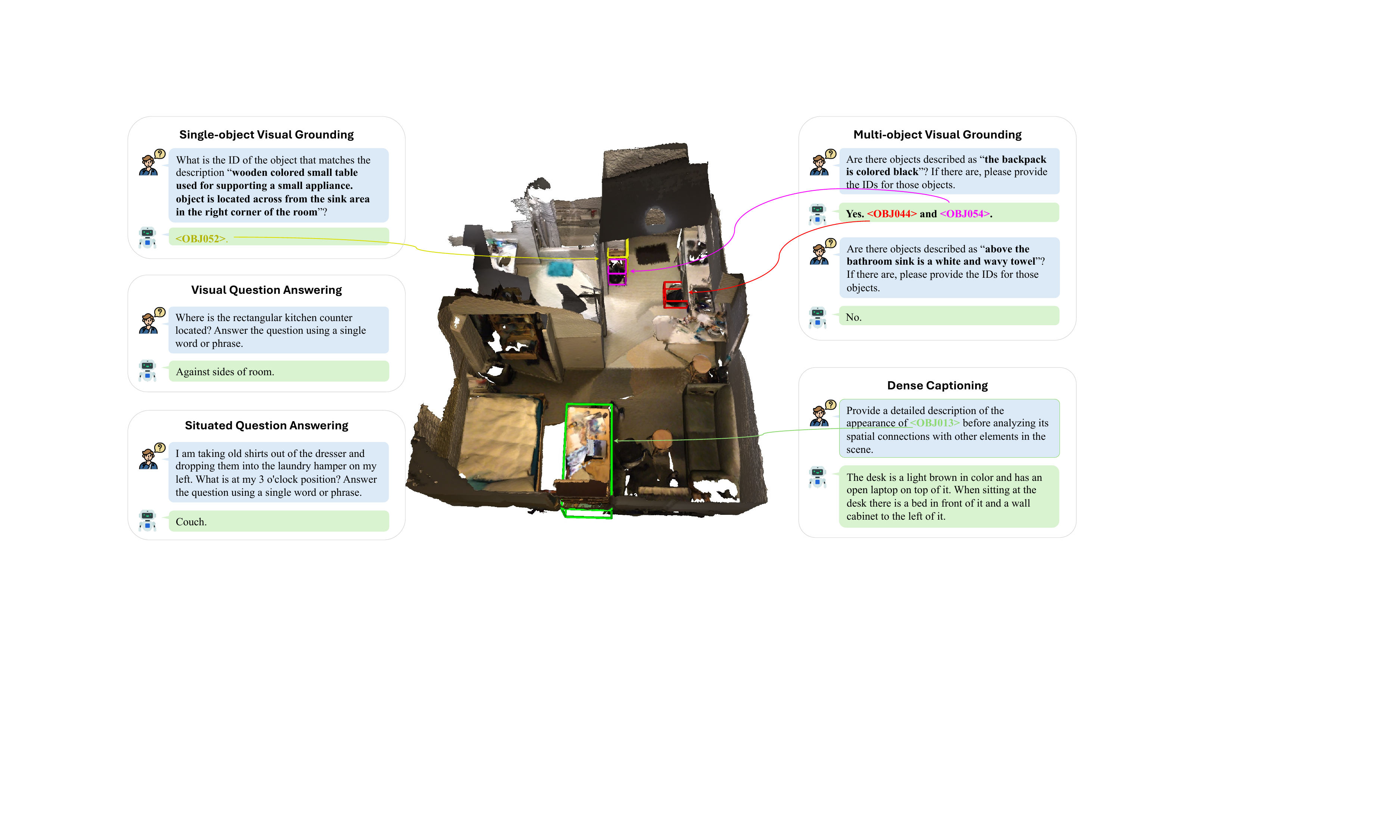}
    \caption{\textbf{Examples of various 3D scene-language understanding tasks}. All the tasks are unified to single-turn question-answering pairs without extra task heads. Object identifiers are used to reference and ground the object during the conversation.}
    \label{fig:taskflow}
\end{figure*}

\noindent\textbf{Training Data}
We aggregate essential training data for downstream tasks and standardize it into uniform instruction formats. The downstream tasks encompass 3D visual grounding (ScanRefer \& Multi3DRef), 3D dense captioning (Scan2Cap), and 3D visual question answering (ScanQA \& SQA3D). We incorporate the training sets from these datasets into our training corpus. Each task is adaptable to a single-turn user-assistant interaction, as illustrated in Figure~\ref{fig:taskflow}.

\noindent\textbf{Training Objective\label{sec:training_and_inference}}
We have unified all tasks into a consistent user-assistant interaction format, and as a result, the sole training loss in the joint-training phase is the Cross-Entropy loss of the language model. The training objective is to optimize the trainable parameters, denoted by $\theta$, aiming to minimize the negative log-likelihood of the target response sequence $s^\mathrm{res}$ generated by the assistant. Specifically, given the input prefix sequence $s^\mathrm{prefix}$, which encompasses both system messages and user instructions, the loss function is expressed as follows:
\begin{equation}
\mathcal{L}(\theta)=-\sum_{i=1}^{k}\log P(s^\mathrm{res}_{i}|s^\mathrm{res}_{[1, \ldots, i-1]}, s^\mathrm{prefix}),
\end{equation}
where $k$ is the number of tokens in the response sequence, and $s^\mathrm{res}_{[1, \ldots, i-1]}$ denotes the sequence of the previous $i-1$ tokens in the response. The set of trainable parameters $\theta$ includes two vision-language projectors, newly added $n$ token embeddings for object identifiers, and the language model itself.

\section{Experiments\label{sec:experiment}}

\subsection{Datasets and Metrics\label{sec:datasets}}

\noindent \textbf{Datasets.}
We conducted experiments on five benchmarks: ScanRefer~\cite{scanrefer} for single-object visual grounding, Multi3DRefer~\cite{multi3drefer} for multi-object visual grounding, Scan2Cap~\cite{scan2cap} for dense captioning, and both ScanQA~\cite{scanqa} and SQA3D~\cite{sqa3d} for visual question answering. These benchmarks are based on the ScanNet dataset~\cite{scannet}, which comprises richly annotated RGB-D scans of real-world indoor scenes, including both 2D and 3D data across 1,513 scans. All benchmarks adhere to the same train/validation/test splits, facilitating joint training and evaluation.

\noindent \textbf{Metrics.}
We adhere to the commonly used metrics in these benchmarks. For ScanRefer~\cite{scanrefer}, we assess thresholded accuracy with Acc@0.25 and Acc@0.5, where predictions are deemed positive if they exhibit higher Intersection over Union (IoU) with the ground truths than the thresholds of 0.25 and 0.5, respectively. In evaluating grounding for a flexible number of target objects in Multi3DRefer~\cite{multi3drefer}, we employ the F1 score at IoU thresholds of 0.25 and 0.5. For Scan2Cap~\cite{scan2cap}, we utilize CIDEr@0.5 and BLEU-4@0.5 (abbreviated as C@0.5 and B-4@0.5), integrating image captioning metrics with the IoU scores between predicted and target bounding boxes. For ScanQA~\cite{scanqa}, the metrics CIDEr~\cite{cider} and BLEU-4~\cite{bleu}, abbreviated as C and B-4, are used. SQA3D~\cite{sqa3d} is evaluated using exact match accuracy (EM) and its refined version, EM-R, as proposed by LEO~\cite{leo}.

\begin{table}[t]
\caption{\textbf{Performance comparison.} ``Expert models'' are tailored for specific tasks using task-oriented heads, while ``LLM-based models'' are designed for general instructions and responses.} \label{tab:performance_comparison}
\resizebox{\linewidth}{!}{
\begin{tabular}{c|c|cccccccccc}
\toprule
 &
  \multirow{2}{*}{Method} &
  \multicolumn{2}{c}{ScanRefer} &
  \multicolumn{2}{c}{Multi3DRefer} &
  \multicolumn{2}{c}{Scan2Cap} &
  \multicolumn{2}{c}{ScanQA} &
  \multicolumn{2}{c}{SQA3D} \\
 &                 & Acc@0.25 & Acc@0.5 & F1@0.25 & F1@0.5 & C@0.5 & B-4@0.5 & C & B-4 & EM   & EM-R \\ \midrule
\multirow{8}{*}{\rotatebox[origin=c]{90}{Expert Models}}   & ScanRefer~\cite{scanrefer}        & 37.3     & 24.3    & -       & -      & -         & -          & -     & -      & -    & -          \\
 & ScanQA~\cite{scanqa}           & -        & -       & -       & -      & -         & -          & 64.9  & 10.1   & -    & -          \\
 & 3DJCG~\cite{3djcg}            & 49.6     & 37.3    & -       & -      & 49.5      & 31.0       & -     & -      & -    & -          \\
 & 3D-VLP~\cite{3d-vlp}           & 51.4     & 39.5    & -       & -      & 54.9      & 32.3       & 67.0  & 11.1   & -    & -          \\
 & M3DRef-CLIP~\cite{multi3drefer}      & 51.9     & 44.7    & 42.8    & 38.4   & -         & -          & -     & -      & -    & -          \\
 & 3D-VisTA~\cite{3dvista}         & 50.6     & 45.5    & -       & -      & 66.9      & 34.0       & 72.9  & 13.1   & 48.5 & -          \\
 & ConcreteNet~\cite{concretenet}      & 50.6     & 46.5    & -       & -      & -         & -          & -     & -      & -    & -          \\
 & Vote2Cap-DETR++~\cite{vote2cap++}  & -        & -       & -       & -      & 67.6      & \textbf{37.1}       & -     & -      & -    & -          \\ \hline
 \\ [-2ex]
 \multirow{7}{*}{\rotatebox[origin=c]{90}{LLM-based Models}} & LAMM~\cite{lamm} & - & 3.4 & - & - & - & - & 42.4 & 5.8 & - & - \\
 & Chat-3D~\cite{chat3d}           & -        & -       & -       & -      & -         & -          & 53.2  & 6.4    & -    & -          \\
 & 3D-LLM~\cite{3dllm}           & 30.3     & -       & -       & -      & -         & -          & 69.4  & 12.0   & -    & -          \\
 & LL3DA~\cite{ll3da}            & -        & -       & -       & -      & 65.2      & 36.8       & 76.8  & 13.5   & -    & -          \\ 
 & LEO~\cite{leo}               & -        & -       & -       & -      & 68.4      & 36.9       & 80.0  & 11.5   & -    & 53.7       \\
 & Scene-LLM~\cite{scenellm}        & -        & -       & -       & -      & -         & -          & 80.0  & 12.0   & 54.2 & -          \\
 & Chat-Scene (Ours)             & \textbf{55.5}     & \textbf{50.2}    & \textbf{57.1}    & \textbf{52.4}   & \textbf{77.1}      & 36.3       & \textbf{87.7}  & \textbf{14.3}   & \textbf{54.6} & \textbf{57.5}     \\ \bottomrule 
\end{tabular}}
\end{table}

\subsection{Comparison with State-of-the-art Methods}

\noindent \textbf{Compared Baselines.} 
The baseline models can be classified into two principal categories: traditional expert models and general multi-modal LLMs. Traditional expert models generate outputs in a fixed format tailored to specific tasks. Conversely, LLM-based models yield open-ended outputs suitable for a broader range of applications.

\begin{itemize}[leftmargin=*]
\item{\textbf{Expert Models:}} Models such as ScanRefer~\cite{scanrefer} and ScanQA~\cite{scanqa} establish initial benchmarks for the ScanRefer and ScanQA datasets, respectively. 3DJCG~\cite{3djcg} integrates multiple tasks within a single architecture, unifying visual grounding and dense captioning tasks due to their synergistic nature. Both 3D-VLP~\cite{3d-vlp} and 3D-VisTA~\cite{3dvista} aim to develop versatile 3D visual-language frameworks by focusing on pre-training strategies for 3D scene-language alignment. M3DRef-CLIP~\cite{multi3drefer} introduces multi-object grounding, enhancing single-object grounding performance. ConcreteNet~\cite{concretenet}, the state-of-the-art (SOTA) model on the ScanRefer benchmark, innovates three methods to augment verbo-visual fusion for dense 3D visual grounding. Vote2Cap-DETR++~\cite{vote2cap++} decouples the processes of caption generation and object localization through parallel decoding, making it the SOTA model on the Scan2Cap benchmark.

\item{\textbf{LLM-based Models:}} LAMM~\cite{lamm} extends research on 2D MLLM to point clouds but lacks a design tailored for 3D scene tasks. Chat-3D~\cite{chat3d} introduces an object-centric method yet fails to address general 3D scene tasks comprehensively. 3D-LLM~\cite{3dllm} utilizes location tokens for object grounding but is constrained by data scarcity. LL3DA~\cite{ll3da} develops an assistant that processes point cloud data directly, responding to textual instructions and visual prompts. LEO~\cite{leo} pioneers an embodied generalist approach by incorporating action tokens. Scene-LLM~\cite{scenellm} merges scene-level and egocentric 3D information, enhancing understanding and reasoning in 3D environments. However, LL3DA, LEO, and Scene-LLM lack grounding capabilities.
\end{itemize}

\noindent \textbf{Analysis.} 
As shown in Table~\ref{tab:performance_comparison}, our model surpasses previous methods across almost all metrics without task-specific fine-tuning, suggesting a promising unified framework for 3D scene understanding. For visual grounding tasks, our model boosts the state-of-the-art performance by 3.7\% (Acc@0.5) on ScanRefer and 14.0\% (F1@0.5) on Multi3DRefer, demonstrating excellent grounding capabilities. For the dense captioning task, we improve the SOTA performance by 8.7\% (CIDEr@0.5) on Scan2Cap, highlighting strong object referring and captioning ability. For VQA tasks on ScanQA and SQA3D, which do not require object referencing and grounding, we still achieve consistent performance enhancement, demonstrating improved overall 3D scene understanding and reasoning.

\begin{table}[tbp]
\caption{\textbf{Ablation studies on object identifiers.} ``Plain Text'' employs plain text for object numbers, ``Gaussian'' uses fixed Gaussian embeddings, and ``Learnable'' learns new identifier tokens. ``Token Cost'' denotes the total tokens for $N$ objects, including object identifiers.} \label{tab:identifier_ablate}
\resizebox{\linewidth}{!}{
\begin{tabular}{ccccccc}
\toprule
\multirow{2}{*}{Identifier Token Type}  & \multirow{2}{*}{Token Cost} & ScanRefer & Multi3DRef & Scan2Cap & ScanQA & SQA3D \\
 &  & Acc@0.5 & F1@0.5 & C@0.5 & CIDEr & EM \\ 
 \midrule
Plain Text & 6$N$ & 47.2 & 49.6 & 73.1 & 84.9 & 53.7 \\ 
Gaussian  & 3$N$ & 46.1 & 49.4 & 71.7 & 82.5 & 53.4 \\
Learnable & 3$N$ & \textbf{50.2} & \textbf{52.4} & \textbf{77.1} & \textbf{87.7} & \textbf{54.6} \\
\bottomrule
\end{tabular}}
\end{table}

\begin{table}[tbp]
\centering
\caption{\textbf{Ablation studies on multi-modal object-centric representations.} ``Early Fusion'' merges object features before language model input, whereas ``Separate Token'' keeps them distinct. ``Single'' denotes using a single image to extract 2D feature of an object, while ``Multi'' uses multi-view images.} \label{tab:feature_ablate}
\resizebox{\linewidth}{!}{
\begin{tabular}{ccccccccc}
\toprule
\multirow{2}{*}{Fusion Method} & \multirow{2}{*}{Token Cost} & \multirow{2}{*}{3D} & \multirow{2}{*}{2D}  & ScanRefer & Multi3DRef & Scan2Cap & ScanQA & SQA3D \\
 &  & & & Acc@0.5 & F1@0.5 & C@0.5 & CIDEr & EM \\ 
 \midrule
 \multirow{3}{*}{--} & \multirow{3}{*}{2$N$} & \cmark & -- & 41.2 & 43.8 & 64.9 & 80.3 & 53.4 \\ 
  &  & \xmark & Single & 32.9 & 37.2 & 65.9 & 83.7 & 53.2 \\ 
  &  & \xmark & Multi & 45.8 & 49.1 & 75.7 & \textbf{88.2} & 54.4 \\
 \midrule
 \multirow{2}{*}{Separate Token} & \multirow{2}{*}{3$N$} & \cmark & Single & 45.7 & 49.1 & 73.8 & 86.5 & 53.2 \\
 & & \cmark & Multi & \textbf{50.2} & \textbf{52.4} & \textbf{77.1} & 87.7 & \textbf{54.6} \\ 
  \midrule
 \multirow{2}{*}{Early Fusion} & \multirow{2}{*}{2$N$} & \cmark & Single & 42.4 & 46.8 & 70.9 & 85.9 & 52.9 \\
 & & \cmark & Multi & 46.9 & 50.2 & 74.4 & 88.0 & 53.9 \\
 \bottomrule
\end{tabular}}
\end{table}

\subsection{Ablation Study}
\label{sec:ablate}
\noindent \textbf{Object Identifiers.}
Table~\ref{tab:identifier_ablate} shows that the format of object identifiers affects both performance and token cost. For comparing token costs, we consider scenes with hundreds of objects. A straightforward approach is to use plain text for object identifiers such as ``Obj001'', which is tokenized into four tokens (``Obj'', ``0'', ``0'', ``1''). Including two object tokens (3D \& 2D), representing a single object requires six tokens in total. This high token cost makes the approach impractical for real-world scenarios. Thus, we explored using a single token per identifier by adding new tokens to the language's vocabulary. We assess two strategies: employing fixed random Gaussian embeddings ("Gaussian") and using learnable tokens ("Learnable"). The results show that learnable tokens enhance performance and reduce token costs simultaneously. Lowering token costs from $6N$ to $3N$ significantly reduces memory usage and accelerates training/inference when handling 3D scenes with numerous objects.

\noindent \textbf{Multi-modal Object-centric Representations.}
We evaluate various methods for retrieving object features and combining features from different sources (3D and 2D), as shown in Table~\ref{tab:feature_ablate}. As described in Section~\ref{sec:model_arch}, the 3D and 2D features are derived from the 3D encoder and 2D encoder, respectively. We assess two methods of extracting 2D features: one from a single-view image (``Single'') and another from multi-view images (``Multi'').

First, we evaluate the performance of using either a single 3D feature or a single 2D feature for the object token. The results show that using a 2D feature derived from multi-view images yields better performance than using a 3D feature. This suggests that semantic information from 2D visual contexts is more crucial than spatial information from 3D point clouds. It may also indicate that the pre-trained 2D encoder is more reliable than the pre-trained 3D encoder due to the abundance of 2D image-text data compared to 3D-text data for pre-training.

Next, we assess the combination of both 3D and 2D features using two fusion methods. The first method, named ``Separate Token'' in the table, involves using two separate object tokens (3D and 2D object tokens) to represent object information. The second method, termed ``Early Fusion'', combines the 3D and 2D features into a single token for each object. The results indicate that combining 3D and 2D features consistently improves performance compared to using a single 3D/2D feature, highlighting the importance of utilizing both modalities. Fusing 3D and 2D tokens reduces the token cost, while it results in a slight performance drop. This provides an option for situations where the token limit is tight, suggesting that combining multi-modal features into one token is acceptable.

\begin{figure*}[tbp]
    \centering
    \includegraphics[width=\textwidth]{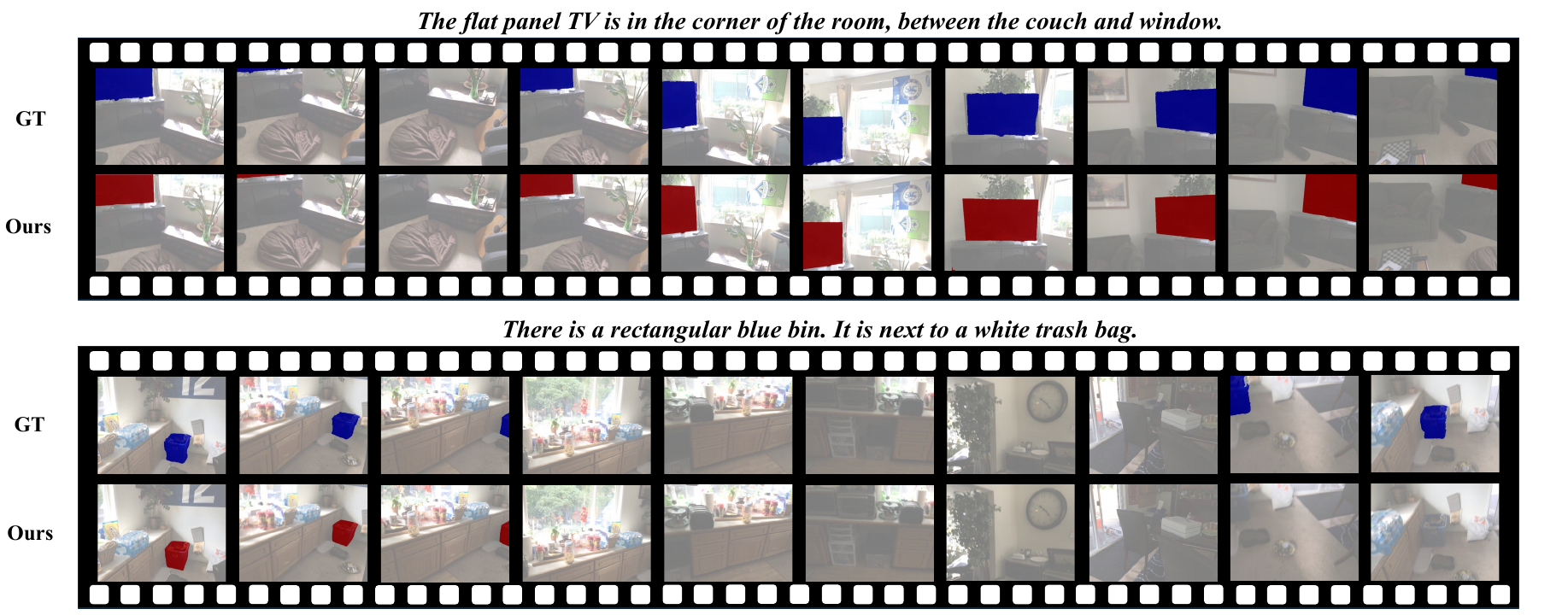}
    \caption{\textbf{Visualization results of video grounding for video input.} ``GT'' denotes the projected 2D masks derived from the ground-truth 3D point cloud mask.}
    \label{fig:video_input_main}
\end{figure*}

\subsection{Experiments with 2D Video Input}

In practical applications of 3D scene understanding, acquiring indoor RGB (video) scan is simpler than obtaining a processed 3D point cloud from RGB-D images. We examine our model's ability to adapt to video input (without depth) for 3D indoor scenes based on the ScanNet~\cite{scannet} dataset. For video input, we use a tracking-based video detector DEVA~\cite{deva} to extract object proposals. This process involves detecting objects in each frame and merging these proposals across frames via the tracking module. After extracting objects from the video, we perform the same operations as for the 3D tasks. The grounding results can then be evaluated on video frames with 2D masks.

\setlength{\intextsep}{2pt}
\begin{wraptable}{R}{0.5\textwidth}
\caption{\textbf{Evaluation results for video input.}} \label{tab:video_input}
\resizebox{\linewidth}{!}{
\begin{tabular}{ccccc}
\toprule
\multirow{2}{*}{Method} & \multicolumn{2}{c}{Video Grounding} & ScanQA & SQA3D \\
 & Acc@0.25 & Acc@0.5 & CIDEr & EM \\ \midrule
Random & 1.4 & 0.5 & -- & -- \\
Ours & \textbf{22.8} & \textbf{10.8} & \textbf{85.6} & \textbf{52.9} \\
{\color{gray}Upperbound} & {\color{gray}54.9} & {\color{gray}22.2} & {\color{gray}--} & {\color{gray}--} \\
\bottomrule
\end{tabular}}
\end{wraptable}

\noindent \textbf{Tasks and Metrics.}
We assess video grounding and VQA tasks for video input. For video grounding, we use descriptions annotated in ScanRefer and project the ground-truth object's point cloud to 2D masks in video frames, allowing us to compute the IoU between the predicted masks and GT masks in 2D images. Given a video with a frame length of $L$, the predicted masks are a series of 2D masks denoted as $\{\mathbf{M}^{\mathrm{p}}_i\in\mathbb{R}^{H \times W} \}_{i=1...L}$ and the GT masks denoted as $\{\mathbf{M}^{\mathrm{g}}_i\in\mathbb{R}^{H \times W} \}_{i=1...L}$, where $H$ and $W$ are the height and width of an image, respectively. We concatenate these masks along the temporal axis to obtain a predicted spatial-temporal mask $\mathbf{\tilde{M}}^\mathrm{p}\in\mathbb{R}^{H \times W \times L}$ and a GT spatial-temporal mask $\mathbf{\tilde{M}}^\mathrm{g}\in\mathbb{R}^{H \times W \times L}$. We propose calculating the Spatial-Temporal IoU (ST-IoU) between the predicted mask $\mathbf{\tilde{M}}^\mathrm{p}$ and the GT mask $\mathbf{\tilde{M}}^\mathrm{g}$. Thus, similar to the metrics for the grounding task on ScanRefer, we use Acc@0.25 and Acc@0.5 to measure accuracy based on the ST-IoU threshold. For VQA tasks, we use the annotations of ScanQA and SQA3D along with their respective metrics for evaluation.

\noindent \textbf{Performance Analysis.}
Table~\ref{tab:video_input} presents the evaluation results of video grounding and VQA tasks. For video grounding, we compute upper bound results to assess the quality of object masks extracted by the video detector. Compared to the upper bound and random results, our method demonstrates strong grounding ability. Visualization results are provided in Figure~\ref{fig:video_input_main}. The second example shows that the tracking-based video detector might lose track of an object after it has been out of sight for a prolonged period. This is a primary reason for the low quality of the extracted object masks. Missing parts of the frames that contain the target object leads to the low Acc@0.5 result of the upper bound. For VQA tasks, we achieve comparable results to those using objects extracted from 3D inputs. This indicates that despite the lower quality of extracted objects from video input, our approach of constructing scene embeddings from sequences of object-level embeddings efficiently enhances overall scene comprehension, thereby improving QA performance on ScanQA and SQA3D.

\section{Conclusion\label{sec:conclusion}}
To enable efficient object referencing and grounding abilities in 3D MLLMs, this paper proposes modeling and interacting with 3D scenes at the object level. It decomposes the input 3D scene into a set of object proposals assigned with object identifiers. Given the scarcity of scene-language data, we model the scene embeddings as a sequence of explicit object-level embeddings, derived from semantic-rich 2D and 3D representations. By using object identifiers, we transform diverse 3D scene-language tasks into a unified question-answering format. With minimal fine-tuning on all downstream tasks, our model significantly outperforms existing methods across various benchmarks.



\newpage
{\small
\bibliographystyle{abbrv}
\bibliography{main}

\begin{thebibliography}{10}

\bibitem{referit3d}
P.~Achlioptas, A.~Abdelreheem, F.~Xia, M.~Elhoseiny, and L.~Guibas.
\newblock Referit3d: Neural listeners for fine-grained 3d object identification in real-world scenes.
\newblock In {\em Computer Vision--ECCV 2020: 16th European Conference, Glasgow, UK, August 23--28, 2020, Proceedings, Part I 16}, pages 422--440. Springer, 2020.

\bibitem{scanqa}
D.~Azuma, T.~Miyanishi, S.~Kurita, and M.~Kawanabe.
\newblock Scanqa: 3d question answering for spatial scene understanding.
\newblock In {\em proceedings of the IEEE/CVF conference on computer vision and pattern recognition}, pages 19129--19139, 2022.

\bibitem{3djcg}
D.~Cai, L.~Zhao, J.~Zhang, L.~Sheng, and D.~Xu.
\newblock 3djcg: A unified framework for joint dense captioning and visual grounding on 3d point clouds.
\newblock In {\em Proceedings of the IEEE/CVF Conference on Computer Vision and Pattern Recognition}, pages 16464--16473, 2022.

\bibitem{scanrefer}
D.~Z. Chen, A.~X. Chang, and M.~Nie{\ss}ner.
\newblock Scanrefer: 3d object localization in rgb-d scans using natural language.
\newblock In {\em European conference on computer vision}, pages 202--221. Springer, 2020.

\bibitem{d3net}
D.~Z. Chen, Q.~Wu, M.~Nie{\ss}ner, and A.~X. Chang.
\newblock D3net: a speaker-listener architecture for semi-supervised dense captioning and visual grounding in rgb-d scans.
\newblock 2021.

\bibitem{ham}
J.~Chen, W.~Luo, X.~Wei, L.~Ma, and W.~Zhang.
\newblock Ham: Hierarchical attention model with high performance for 3d visual grounding.
\newblock {\em arXiv preprint arXiv:2210.12513}, 2, 2022.

\bibitem{ll3da}
S.~Chen, X.~Chen, C.~Zhang, M.~Li, G.~Yu, H.~Fei, H.~Zhu, J.~Fan, and T.~Chen.
\newblock Ll3da: Visual interactive instruction tuning for omni-3d understanding, reasoning, and planning.
\newblock {\em arXiv preprint arXiv:2311.18651}, 2023.

\bibitem{vil3drel}
S.~Chen, P.-L. Guhur, M.~Tapaswi, C.~Schmid, and I.~Laptev.
\newblock Language conditioned spatial relation reasoning for 3d object grounding.
\newblock {\em Advances in Neural Information Processing Systems}, 35:20522--20535, 2022.

\bibitem{vote2cap}
S.~Chen, H.~Zhu, X.~Chen, Y.~Lei, G.~Yu, and T.~Chen.
\newblock End-to-end 3d dense captioning with vote2cap-detr.
\newblock In {\em Proceedings of the IEEE/CVF Conference on Computer Vision and Pattern Recognition}, pages 11124--11133, 2023.

\bibitem{vote2cap++}
S.~Chen, H.~Zhu, M.~Li, X.~Chen, P.~Guo, Y.~Lei, Y.~Gang, T.~Li, and T.~Chen.
\newblock Vote2cap-detr++: Decoupling localization and describing for end-to-end 3d dense captioning.
\newblock {\em IEEE Transactions on Pattern Analysis and Machine Intelligence}, 2024.

\bibitem{focalformer3d}
Y.~Chen, Z.~Yu, Y.~Chen, S.~Lan, A.~Anandkumar, J.~Jia, and J.~M. Alvarez.
\newblock Focalformer3d: focusing on hard instance for 3d object detection.
\newblock In {\em Proceedings of the IEEE/CVF International Conference on Computer Vision}, pages 8394--8405, 2023.

\bibitem{scan2cap}
Z.~Chen, A.~Gholami, M.~Nie{\ss}ner, and A.~X. Chang.
\newblock Scan2cap: Context-aware dense captioning in rgb-d scans.
\newblock In {\em Proceedings of the IEEE/CVF conference on computer vision and pattern recognition}, pages 3193--3203, 2021.

\bibitem{deva}
H.~K. Cheng, S.~W. Oh, B.~Price, A.~Schwing, and J.-Y. Lee.
\newblock Tracking anything with decoupled video segmentation.
\newblock In {\em Proceedings of the IEEE/CVF International Conference on Computer Vision}, pages 1316--1326, 2023.

\bibitem{chiang2023vicuna}
W.-L. Chiang, Z.~Li, Z.~Lin, Y.~Sheng, Z.~Wu, H.~Zhang, L.~Zheng, S.~Zhuang, Y.~Zhuang, J.~E. Gonzalez, et~al.
\newblock Vicuna: An open-source chatbot impressing gpt-4 with 90\%* chatgpt quality.
\newblock {\em See https://vicuna. lmsys. org (accessed 14 April 2023)}, 2023.

\bibitem{chowdhery2022palm}
A.~Chowdhery, S.~Narang, J.~Devlin, M.~Bosma, G.~Mishra, A.~Roberts, P.~Barham, H.~W. Chung, C.~Sutton, S.~Gehrmann, et~al.
\newblock Palm: Scaling language modeling with pathways.
\newblock {\em arXiv preprint arXiv:2204.02311}, 2022.

\bibitem{scannet}
A.~Dai, A.~X. Chang, M.~Savva, M.~Halber, T.~Funkhouser, and M.~Nie{\ss}ner.
\newblock Scannet: Richly-annotated 3d reconstructions of indoor scenes.
\newblock In {\em Proceedings of the IEEE conference on computer vision and pattern recognition}, pages 5828--5839, 2017.

\bibitem{scenellm}
R.~Fu, J.~Liu, X.~Chen, Y.~Nie, and W.~Xiong.
\newblock Scene-llm: Extending language model for 3d visual understanding and reasoning.
\newblock {\em arXiv preprint arXiv:2403.11401}, 2024.

\bibitem{pointbind}
Z.~Guo, R.~Zhang, X.~Zhu, Y.~Tang, X.~Ma, J.~Han, K.~Chen, P.~Gao, X.~Li, H.~Li, et~al.
\newblock Point-bind \& point-llm: Aligning point cloud with multi-modality for 3d understanding, generation, and instruction following.
\newblock {\em arXiv preprint arXiv:2309.00615}, 2023.

\bibitem{han2023imagebind}
J.~Han, R.~Zhang, W.~Shao, P.~Gao, P.~Xu, H.~Xiao, K.~Zhang, C.~Liu, S.~Wen, Z.~Guo, et~al.
\newblock Imagebind-llm: Multi-modality instruction tuning.
\newblock {\em arXiv preprint arXiv:2309.03905}, 2023.

\bibitem{3dllm}
Y.~Hong, H.~Zhen, P.~Chen, S.~Zheng, Y.~Du, Z.~Chen, and C.~Gan.
\newblock 3d-llm: Injecting the 3d world into large language models.
\newblock {\em arXiv preprint arXiv:2307.12981}, 2023.

\bibitem{lora}
E.~J. Hu, Y.~Shen, P.~Wallis, Z.~Allen-Zhu, Y.~Li, S.~Wang, L.~Wang, and W.~Chen.
\newblock Lora: Low-rank adaptation of large language models.
\newblock {\em arXiv preprint arXiv:2106.09685}, 2021.

\bibitem{leo}
J.~Huang, S.~Yong, X.~Ma, X.~Linghu, P.~Li, Y.~Wang, Q.~Li, S.-C. Zhu, B.~Jia, and S.~Huang.
\newblock An embodied generalist agent in 3d world.
\newblock {\em arXiv preprint arXiv:2311.12871}, 2023.

\bibitem{tgnn}
P.-H. Huang, H.-H. Lee, H.-T. Chen, and T.-L. Liu.
\newblock Text-guided graph neural networks for referring 3d instance segmentation.
\newblock In {\em Proceedings of the AAAI Conference on Artificial Intelligence}, volume~35, pages 1610--1618, 2021.

\bibitem{mvt}
S.~Huang, Y.~Chen, J.~Jia, and L.~Wang.
\newblock Multi-view transformer for 3d visual grounding.
\newblock In {\em Proceedings of the IEEE/CVF Conference on Computer Vision and Pattern Recognition}, pages 15524--15533, 2022.

\bibitem{butd-detr}
A.~Jain, N.~Gkanatsios, I.~Mediratta, and K.~Fragkiadaki.
\newblock Bottom up top down detection transformers for language grounding in images and point clouds.
\newblock In {\em European Conference on Computer Vision}, pages 417--433. Springer, 2022.

\bibitem{pointgroup}
L.~Jiang, H.~Zhao, S.~Shi, S.~Liu, C.-W. Fu, and J.~Jia.
\newblock Pointgroup: Dual-set point grouping for 3d instance segmentation.
\newblock In {\em Proceedings of the IEEE/CVF conference on computer vision and Pattern recognition}, pages 4867--4876, 2020.

\bibitem{more}
Y.~Jiao, S.~Chen, Z.~Jie, J.~Chen, L.~Ma, and Y.-G. Jiang.
\newblock More: Multi-order relation mining for dense captioning in 3d scenes.
\newblock In {\em European Conference on Computer Vision}, pages 528--545. Springer, 2022.

\bibitem{3d-vlp}
Z.~Jin, M.~Hayat, Y.~Yang, Y.~Guo, and Y.~Lei.
\newblock Context-aware alignment and mutual masking for 3d-language pre-training.
\newblock In {\em Proceedings of the IEEE/CVF Conference on Computer Vision and Pattern Recognition}, pages 10984--10994, 2023.

\bibitem{oneformer3d}
M.~Kolodiazhnyi, A.~Vorontsova, A.~Konushin, and D.~Rukhovich.
\newblock Oneformer3d: One transformer for unified point cloud segmentation.
\newblock {\em arXiv preprint arXiv:2311.14405}, 2023.

\bibitem{lisa}
X.~Lai, Z.~Tian, Y.~Chen, Y.~Li, Y.~Yuan, S.~Liu, and J.~Jia.
\newblock Lisa: Reasoning segmentation via large language model.
\newblock {\em arXiv preprint arXiv:2308.00692}, 2023.

\bibitem{li2023otter}
B.~Li, Y.~Zhang, L.~Chen, J.~Wang, J.~Yang, and Z.~Liu.
\newblock Otter: A multi-modal model with in-context instruction tuning.
\newblock {\em arXiv preprint arXiv:2305.03726}, 2023.

\bibitem{li2023videochat}
K.~Li, Y.~He, Y.~Wang, Y.~Li, W.~Wang, P.~Luo, Y.~Wang, L.~Wang, and Y.~Qiao.
\newblock Videochat: Chat-centric video understanding.
\newblock {\em arXiv preprint arXiv:2305.06355}, 2023.

\bibitem{llava1-5}
H.~Liu, C.~Li, Y.~Li, and Y.~J. Lee.
\newblock Improved baselines with visual instruction tuning.
\newblock {\em arXiv preprint arXiv:2310.03744}, 2023.

\bibitem{llava}
H.~Liu, C.~Li, Q.~Wu, and Y.~J. Lee.
\newblock Visual instruction tuning.
\newblock {\em arXiv preprint arXiv:2304.08485}, 2023.

\bibitem{3d-sps}
J.~Luo, J.~Fu, X.~Kong, C.~Gao, H.~Ren, H.~Shen, H.~Xia, and S.~Liu.
\newblock 3d-sps: Single-stage 3d visual grounding via referred point progressive selection.
\newblock In {\em Proceedings of the IEEE/CVF Conference on Computer Vision and Pattern Recognition}, pages 16454--16463, 2022.

\bibitem{sqa3d}
X.~Ma, S.~Yong, Z.~Zheng, Q.~Li, Y.~Liang, S.-C. Zhu, and S.~Huang.
\newblock Sqa3d: Situated question answering in 3d scenes.
\newblock {\em arXiv preprint arXiv:2210.07474}, 2022.

\bibitem{gpt4}
R.~OpenAI.
\newblock Gpt-4 technical report. arxiv 2303.08774.
\newblock {\em View in Article}, 2, 2023.

\bibitem{dinov2}
M.~Oquab, T.~Darcet, T.~Moutakanni, H.~Vo, M.~Szafraniec, V.~Khalidov, P.~Fernandez, D.~Haziza, F.~Massa, A.~El-Nouby, et~al.
\newblock Dinov2: Learning robust visual features without supervision.
\newblock {\em arXiv preprint arXiv:2304.07193}, 2023.

\bibitem{bleu}
K.~Papineni, S.~Roukos, T.~Ward, and W.-J. Zhu.
\newblock Bleu: a method for automatic evaluation of machine translation.
\newblock In {\em Proceedings of the 40th annual meeting of the Association for Computational Linguistics}, pages 311--318, 2002.

\bibitem{parelli2023clip}
M.~Parelli, A.~Delitzas, N.~Hars, G.~Vlassis, S.~Anagnostidis, G.~Bachmann, and T.~Hofmann.
\newblock Clip-guided vision-language pre-training for question answering in 3d scenes.
\newblock In {\em Proceedings of the IEEE/CVF Conference on Computer Vision and Pattern Recognition}, pages 5606--5611, 2023.

\bibitem{mask3d}
J.~Schult, F.~Engelmann, A.~Hermans, O.~Litany, S.~Tang, and B.~Leibe.
\newblock Mask3d: Mask transformer for 3d semantic instance segmentation.
\newblock In {\em 2023 IEEE International Conference on Robotics and Automation (ICRA)}, pages 8216--8223. IEEE, 2023.

\bibitem{v-detr}
Y.~Shen, Z.~Geng, Y.~Yuan, Y.~Lin, Z.~Liu, C.~Wang, H.~Hu, N.~Zheng, and B.~Guo.
\newblock V-detr: Detr with vertex relative position encoding for 3d object detection.
\newblock {\em arXiv preprint arXiv:2308.04409}, 2023.

\bibitem{touvron2023llama}
H.~Touvron, T.~Lavril, G.~Izacard, X.~Martinet, M.-A. Lachaux, T.~Lacroix, B.~Rozi{\`e}re, N.~Goyal, E.~Hambro, F.~Azhar, et~al.
\newblock Llama: Open and efficient foundation language models.
\newblock {\em arXiv preprint arXiv:2302.13971}, 2023.

\bibitem{concretenet}
O.~Unal, C.~Sakaridis, S.~Saha, F.~Yu, and L.~Van~Gool.
\newblock Three ways to improve verbo-visual fusion for dense 3d visual grounding.
\newblock {\em arXiv preprint arXiv:2309.04561}, 2023.

\bibitem{cider}
R.~Vedantam, C.~Lawrence~Zitnick, and D.~Parikh.
\newblock Cider: Consensus-based image description evaluation.
\newblock In {\em Proceedings of the IEEE conference on computer vision and pattern recognition}, pages 4566--4575, 2015.

\bibitem{softgroup}
T.~Vu, K.~Kim, T.~M. Luu, T.~Nguyen, and C.~D. Yoo.
\newblock Softgroup for 3d instance segmentation on point clouds.
\newblock In {\em Proceedings of the IEEE/CVF Conference on Computer Vision and Pattern Recognition}, pages 2708--2717, 2022.

\bibitem{3drp-net}
Z.~Wang, H.~Huang, Y.~Zhao, L.~Li, X.~Cheng, Y.~Zhu, A.~Yin, and Z.~Zhao.
\newblock 3drp-net: 3d relative position-aware network for 3d visual grounding.
\newblock {\em arXiv preprint arXiv:2307.13363}, 2023.

\bibitem{wang2023distilling}
Z.~Wang, H.~Huang, Y.~Zhao, L.~Li, X.~Cheng, Y.~Zhu, A.~Yin, and Z.~Zhao.
\newblock Distilling coarse-to-fine semantic matching knowledge for weakly supervised 3d visual grounding.
\newblock In {\em Proceedings of the IEEE/CVF International Conference on Computer Vision}, pages 2662--2671, 2023.

\bibitem{chat3d}
Z.~Wang, H.~Huang, Y.~Zhao, Z.~Zhang, and Z.~Zhao.
\newblock Chat-3d: Data-efficiently tuning large language model for universal dialogue of 3d scenes.
\newblock {\em arXiv preprint arXiv:2308.08769}, 2023.

\bibitem{dora}
T.-Y. Wu, S.-Y. Huang, and Y.-C.~F. Wang.
\newblock Dora: 3d visual grounding with order-aware referring.
\newblock {\em arXiv preprint arXiv:2403.16539}, 2024.

\bibitem{eda}
Y.~Wu, X.~Cheng, R.~Zhang, Z.~Cheng, and J.~Zhang.
\newblock Eda: Explicit text-decoupling and dense alignment for 3d visual grounding.
\newblock In {\em Proceedings of the IEEE/CVF Conference on Computer Vision and Pattern Recognition}, pages 19231--19242, 2023.

\bibitem{pointllm}
R.~Xu, X.~Wang, T.~Wang, Y.~Chen, J.~Pang, and D.~Lin.
\newblock Pointllm: Empowering large language models to understand point clouds.
\newblock {\em arXiv preprint arXiv:2308.16911}, 2023.

\bibitem{llmgrounder}
J.~Yang, X.~Chen, S.~Qian, N.~Madaan, M.~Iyengar, D.~F. Fouhey, and J.~Chai.
\newblock Llm-grounder: Open-vocabulary 3d visual grounding with large language model as an agent.
\newblock {\em arXiv preprint arXiv:2309.12311}, 2023.

\bibitem{sat}
Z.~Yang, S.~Zhang, L.~Wang, and J.~Luo.
\newblock Sat: 2d semantics assisted training for 3d visual grounding.
\newblock In {\em Proceedings of the IEEE/CVF International Conference on Computer Vision}, pages 1856--1866, 2021.

\bibitem{ye2023mplug}
Q.~Ye, H.~Xu, G.~Xu, J.~Ye, M.~Yan, Y.~Zhou, J.~Wang, A.~Hu, P.~Shi, Y.~Shi, et~al.
\newblock mplug-owl: Modularization empowers large language models with multimodality.
\newblock {\em arXiv preprint arXiv:2304.14178}, 2023.

\bibitem{lamm}
Z.~Yin, J.~Wang, J.~Cao, Z.~Shi, D.~Liu, M.~Li, X.~Huang, Z.~Wang, L.~Sheng, L.~Bai, et~al.
\newblock Lamm: Language-assisted multi-modal instruction-tuning dataset, framework, and benchmark.
\newblock {\em Advances in Neural Information Processing Systems}, 36, 2024.

\bibitem{ferret}
H.~You, H.~Zhang, Z.~Gan, X.~Du, B.~Zhang, Z.~Wang, L.~Cao, S.-F. Chang, and Y.~Yang.
\newblock Ferret: Refer and ground anything anywhere at any granularity.
\newblock {\em arXiv preprint arXiv:2310.07704}, 2023.

\bibitem{X-trans2cap}
Z.~Yuan, X.~Yan, Y.~Liao, Y.~Guo, G.~Li, S.~Cui, and Z.~Li.
\newblock X-trans2cap: Cross-modal knowledge transfer using transformer for 3d dense captioning.
\newblock In {\em Proceedings of the IEEE/CVF Conference on Computer Vision and Pattern Recognition}, pages 8563--8573, 2022.

\bibitem{instancerefer}
Z.~Yuan, X.~Yan, Y.~Liao, R.~Zhang, S.~Wang, Z.~Li, and S.~Cui.
\newblock Instancerefer: Cooperative holistic understanding for visual grounding on point clouds through instance multi-level contextual referring.
\newblock In {\em Proceedings of the IEEE/CVF International Conference on Computer Vision}, pages 1791--1800, 2021.

\bibitem{sam3d}
D.~Zhang, D.~Liang, H.~Yang, Z.~Zou, X.~Ye, Z.~Liu, and X.~Bai.
\newblock Sam3d: Zero-shot 3d object detection via segment anything model.
\newblock {\em arXiv preprint arXiv:2306.02245}, 2023.

\bibitem{ferretv2}
H.~Zhang, H.~You, P.~Dufter, B.~Zhang, C.~Chen, H.-Y. Chen, T.-J. Fu, W.~Y. Wang, S.-F. Chang, Z.~Gan, et~al.
\newblock Ferret-v2: An improved baseline for referring and grounding with large language models.
\newblock {\em arXiv preprint arXiv:2404.07973}, 2024.

\bibitem{multi3drefer}
Y.~Zhang, Z.~Gong, and A.~X. Chang.
\newblock Multi3drefer: Grounding text description to multiple 3d objects.
\newblock In {\em Proceedings of the IEEE/CVF International Conference on Computer Vision}, pages 15225--15236, 2023.

\bibitem{3dvg-transformer}
L.~Zhao, D.~Cai, L.~Sheng, and D.~Xu.
\newblock 3dvg-transformer: Relation modeling for visual grounding on point clouds.
\newblock In {\em Proceedings of the IEEE/CVF International Conference on Computer Vision}, pages 2928--2937, 2021.

\bibitem{zhao2023bubogpt}
Y.~Zhao, Z.~Lin, D.~Zhou, Z.~Huang, J.~Feng, and B.~Kang.
\newblock Bubogpt: Enabling visual grounding in multi-modal llms.
\newblock {\em arXiv preprint arXiv:2307.08581}, 2023.

\bibitem{uni3d}
J.~Zhou, J.~Wang, B.~Ma, Y.-S. Liu, T.~Huang, and X.~Wang.
\newblock Uni3d: Exploring unified 3d representation at scale.
\newblock {\em arXiv preprint arXiv:2310.06773}, 2023.

\bibitem{zhu2023minigpt}
D.~Zhu, J.~Chen, X.~Shen, X.~Li, and M.~Elhoseiny.
\newblock Minigpt-4: Enhancing vision-language understanding with advanced large language models.
\newblock {\em arXiv preprint arXiv:2304.10592}, 2023.

\bibitem{3dvista}
Z.~Zhu, X.~Ma, Y.~Chen, Z.~Deng, S.~Huang, and Q.~Li.
\newblock 3d-vista: Pre-trained transformer for 3d vision and text alignment.
\newblock In {\em Proceedings of the IEEE/CVF International Conference on Computer Vision}, pages 2911--2921, 2023.

\end{thebibliography}
}


\newpage
\appendix

\section{Implementation Details\label{appendix:implementation}}
For the 3D point cloud input, we utilize the 3D instance segmentation model Mask3D~\cite{mask3d} to extract 100 object proposals. We then employ the pre-trained encoder Uni3D~\cite{uni3d} to obtain 3D features and DINOv2~\cite{dinov2} for 2D features. Both the 3D-language projector $f_p(\cdot)$ and the 2D-language projector $f_v(\cdot)$ are three-layer MLPs. For the video input, we use the open-vocabulary video segmentation model DEVA~\cite{deva} to extract object proposals, with an average object number of 48. We choose the Vicuna-7B-v1.5 model~\cite{chiang2023vicuna} as the language model, which is based on LLaMA 2~\cite{touvron2023llama}. We fine-tune the language model using LoRA~\cite{lora}, with the rank set to 16. The base learning rate is set to 5e-6 with a cosine annealing schedule, and the batch size is 32. The training takes 2 epochs and the total training step is 3200. The entire training process takes approximately 8 hours on 4 NVIDIA A100 GPUs.

\noindent \textbf{2D Encoder.}
We describe the method for deriving 2D representations from the projected input masks and multi-view images. The 2D model DINOv2~\cite{dinov2} computes a 257$\times$1024 feature vector for each image, comprising a 1$\times$1024 CLS feature and 256$\times$1024 patch features. These patch features represent local areas in the images divided into 16$\times$16 patches. For each object mask within an image, we extract patch features only where the patch intersects with the mask. We then average these extracted patch features along the patch axis to obtain a 1$\times$1024 feature vector per image. For patch features sourced from multi-view images, we average them weighted by the mask size on each image. We apply this procedure to extract DINOv2 features from multi-view images for each object, thus generating the final object representations. In our ablation study, described in Section~\ref{sec:ablate}, we introduce a ``Single'' approach where we select the image with the largest mask for the current object and extract features solely from this image.

\section{Quantitative Results}

For the five datasets, we employ evaluation metrics as proposed in their respective original publications. We compare our model against a comprehensive array of state-of-the-art methods, as detailed in Table~\ref{tab:scanrefer} for ScanRefer~\cite{scanrefer}, Table~\ref{tab:multi3drefer} for Multi3DRefer~\cite{multi3drefer}, Table~\ref{tab:scan2cap} for Scan2Cap~\cite{scan2cap}, Table~\ref{tab:scanqa} for ScanQA~\cite{scanqa}, and Table~\ref{tab:sqa3d} for SQA3D~\cite{sqa3d}.

\begin{table}[tp]
\caption{\textbf{Performance comparison on the validation set of ScanRefer~\cite{scanrefer}.}} \label{tab:scanrefer}
\resizebox{\linewidth}{!}{
\begin{tabular}{cccccccc}
\toprule
\multirow{2}{*}{Method} & \multirow{2}{*}{Venue} & \multicolumn{2}{c}{Unique} & \multicolumn{2}{c}{Multiple} & \multicolumn{2}{c}{Overall} \\
 &  & Acc@0.25 & Acc@0.5 & Acc@0.25 & Acc@0.5 & Acc@0.25 & Acc@0.5 \\
 \midrule
ScanRefer~\cite{scanrefer} & ECCV20 & 76.33 & 53.51 & 32.73 & 21.11 & 41.19 & 27.40 \\
TGNN~\cite{tgnn} & AAAI21 & 68.61 & 56.80 & 29.84 & 23.18 & 37.37 & 29.70 \\
SAT~\cite{sat} & ICCV21 & 73.21 & 50.83 & 37.64 & 25.16 & 44.54 & 30.14 \\
InstanceRefer~\cite{instancerefer} & ICCV21 & 75.72 & 64.66 & 29.41 & 22.99 & 38.40 & 31.08 \\
3DVG-Transformer~\cite{3dvg-transformer} & ICCV21 & 81.93 & 60.64 & 39.30 & 28.42 & 47.57 & 34.67 \\
MVT~\cite{mvt} & CVPR22 & 77.67 & 66.45 & 31.92 & 25.26 & 40.80 & 33.26 \\
3D-SPS~\cite{3d-sps} & CVPR22 & 84.12 & 66.72 & 40.32 & 29.82 & 48.82 & 36.98 \\
ViL3DRel~\cite{vil3drel} & NeurIPS22 & 81.58 & 68.62 & 40.30 & 30.71 & 47.94 & 37.73 \\
3DJCG~\cite{3djcg} & CVPR22 & 83.47 & 64.34 & 41.39 & 30.82 & 49.56 & 37.33 \\
D3Net~\cite{d3net} & ECCV22 & -- & 72.04 & -- & 30.05 & -- & 37.87 \\
BUTD-DETR~\cite{butd-detr} & ECCV22 & 84.2 & 66.3 & 46.6 & 35.1 & 52.2 & 39.8 \\
HAM~\cite{ham} & ArXiv22 & 79.24 & 67.86 & 41.46 & 34.03 & 48.79 & 40.60 \\
3DRP-Net~\cite{3drp-net} & EMNLP23 & 83.13 & 67.74 & 42.14 & 31.95 & 50.10 & 38.90 \\
3D-VLP~\cite{3d-vlp} & CVPR23 & 84.23 & 64.61 & 43.51 & 33.41 & 51.41 & 39.46 \\
EDA~\cite{eda} & CVPR23 & 85.76 & 68.57 & \textbf{49.13} & 37.64 & 54.59 & 42.26 \\
M3DRef-CLIP~\cite{multi3drefer} & ICCV23 & 85.3 & 77.2 & 43.8 & 36.8 & 51.9 & 44.7 \\
3D-VisTA~\cite{3dvista} & ICCV23 & 81.6 & 75.1 & 43.7 & 39.1 & 50.6 & 45.8 \\
ConcreteNet~\cite{concretenet} & ECCV24 & 86.40 & 82.05 & 42.41 & 38.39 & 50.61 & 46.53 \\
DOrA~\cite{dora} & ArXiv24 & -- & -- & -- & -- & 52.80 & 44.80 \\
\textbf{Ours} & -- & \textbf{89.59} & \textbf{82.49} & 47.78 & \textbf{42.90} & \textbf{55.52} & \textbf{50.23} \\
\bottomrule
\end{tabular}}
\end{table}

\begin{table}[tp]
\caption{\textbf{Performance comparison on the validation set of Multi3DRefer~\cite{multi3drefer}.}} \label{tab:multi3drefer}
\resizebox{\linewidth}{!}{
\begin{tabular}{cccccccccccc}
\toprule
\multirow{2}{*}{Method} & \multirow{2}{*}{Venue} & ZT w/o D & ZT w/ D & \multicolumn{2}{c}{ST w/o D} & \multicolumn{2}{c}{ST w/ D} & \multicolumn{2}{c}{MT} & \multicolumn{2}{c}{ALL} \\
 &  & F1 & F1 & F1@0.25 & F1@0.5 & F1@0.25 & F1@0.5 & F1@0.25 & F1@0.5 & F1@0.25 & F1@0.5 \\
 \midrule
3DVG-Trans+~\cite{3dvg-transformer} & ICCV21 & 87.1 & 45.8 & -- & 27.5 & -- & 16.7 & -- & 26.5 & -- & 25.5 \\
D3Net (Grounding)~\cite{d3net} & ECCV22 & 81.6 & 32.5 & -- & 38.6 & -- & 23.3 & -- & 35.0 & -- & 32.2 \\
3DJCG (Grounding)~\cite{3djcg} & CVPR22 & \textbf{94.1} & \textbf{66.9} & -- & 26.0 & -- & 16.7 & -- & 26.2 & -- & 26.6 \\
M3DRef-CLIP~\cite{multi3drefer} & ICCV23 & 81.8 & 39.4 & 53.5 & 47.8 & 34.6 & 30.6 & 43.6 & 37.9 & 42.8 & 38.4 \\
\textbf{Ours} & -- & 90.3 & 62.6 & \textbf{82.9} & \textbf{75.9} & \textbf{49.1} & \textbf{44.5} & \textbf{45.7} & \textbf{41.1} & \textbf{57.1} & \textbf{52.4} \\
\bottomrule
\end{tabular}}
\end{table}

\begin{table}[tp]
\caption{\textbf{Performance comparison on the validation set of Scan2Cap~\cite{scan2cap}.}} \label{tab:scan2cap}
\resizebox{\linewidth}{!}{
\begin{tabular}{cccccccccc}
\toprule
\multirow{2}{*}{Method} & \multirow{2}{*}{Venue} & \multicolumn{4}{c}{@0.25} & \multicolumn{4}{c}{@0.5} \\
 & & C & B-4 & M & R & C & B-4 & M & R \\
\midrule
Scan2Cap~\cite{scan2cap} & CVPR21 & 56.82 & 34.18 & 26.29 & 55.27 & 39.08 & 23.32 & 21.97 & 44.48 \\
3DJCG~\cite{3djcg} & CVPR22 & 64.70 & 40.17 & 27.66 & 59.23 & 49.48 & 31.03 & 24.22 & 50.80 \\
X-Trans2Cap~\cite{X-trans2cap} & CVPR22 & 61.83 & 35.65 & 26.61 & 54.70 & 43.87 & 25.05 & 22.46 & 45.28 \\
D3Net~\cite{d3net} & ECCV22 & -- & -- & -- & -- & 62.64 & 35.68 & 25.72 & 53.90 \\
3D-VLP~\cite{3d-vlp} & CVPR23 & 70.73 & 41.03 & 28.14 & 59.72 & 54.94 & 32.31 & 24.83 & 51.51 \\
Vote2Cap-DETR~\cite{vote2cap} & CVPR23 & 71.45 & 39.34 & 28.25 & 59.33 & 61.81 & 34.46 & 26.22 & 54.40 \\
3D-VisTA & ICCV23 & 71.0 & 36.5 & 28.4 & 57.6 & 66.9 & 34.0 & 27.1 & 54.3 \\
LL3DA & CVPR24 & 74.17 & \textbf{41.41} & 27.76 & 59.53 & 65.19 & 36.79 & 25.97 & 55.06 \\
LEO & ICML24 & -- & -- & -- & -- & 68.4 & 36.9 & 27.7 & 57.8 \\
Vote2Cap-DETR++~\cite{vote2cap++} & T-PAMI24 & 76.36 & 41.37 & 28.70 & 60.00 & 67.58 & \textbf{37.05} & 26.89 & 55.64 \\
\textbf{Ours} & -- & \textbf{81.94} & 38.23 & \textbf{29.01} & \textbf{60.57} & \textbf{77.19} & 36.34 & \textbf{28.01} & \textbf{58.12} \\
\bottomrule
\end{tabular}}
\end{table}

\begin{table}[tp]
\caption{\textbf{Performance comparison on the validation set of ScanQA~\cite{scanqa}.}} \label{tab:scanqa}
\resizebox{\linewidth}{!}{
\begin{tabular}{ccccccccccc}
\toprule
Method & Venue & EM@1 & B-1 & B-2 & B-3 & B-4 & ROUGE-L & METEOR & CIDEr & SPICE \\
\midrule
ScanQA~\cite{scanqa} & CVPR22 & 21.05 & 30.24 & 20.40 & 15.11 & 10.08 & 33.33 & 13.14 & 64.86 & 13.43 \\
3D-VLP~\cite{3d-vlp} & CVPR23 & 21.65 & 30.53 & 21.33 & 16.67 & 11.15 & 34.51 & 13.53 & 66.97 & 14.18 \\
3D-LLM~\cite{3dllm} & NeurIPS23 & 20.5 & 39.3 & 25.2 & 18.4 & 12.0 & 35.7 & 14.5 & 69.4 & -- \\
LL3DA~\cite{ll3da} & CVPR24 & -- & -- & -- & \multicolumn{1}{l}{--} & 13.53 & 37.31 & 15.88 & 76.79 & -- \\
LEO~\cite{leo} & ICML24 & -- & -- & -- & -- & 11.5 & 39.3 & 16.2 & 80.0 & -- \\
Scene-LLM~\cite{scenellm} & ArXiv24 & \textbf{27.2} & \textbf{43.6} & 26.8 & 19.1 & 12.0 & 40.0 & 16.6 & 80.0 & -- \\
\textbf{Ours} & -- & 21.62 & 43.20 & \textbf{29.06} & \textbf{20.57} & \textbf{14.31} & \textbf{41.56} & \textbf{18.00} & \textbf{87.70} & \textbf{20.44} \\
\bottomrule
\end{tabular}}
\end{table}

\begin{table}[tp]
\caption{\textbf{Performance comparison on the test set of SQA3D~\cite{sqa3d}.}} \label{tab:sqa3d}
\resizebox{\linewidth}{!}{
\begin{tabular}{ccccccccc}
\toprule
\multirow{2}{*}{Method} & \multirow{2}{*}{Venue} & \multicolumn{6}{c}{Test set} & \multirow{2}{*}{Avg.} \\ \cmidrule{3-8}
 & & What & Is & How & Can & Which & Others &  \\
\midrule
SQA3D~\cite{sqa3d} & ICLR23 & 31.64 & 63.80 & 46.02 & 69.53 & 43.87 & 45.34 & 46.58 \\
3D-VisTA~\cite{3dvista} & ICCV23 & 34.8 & 63.3 & 45.4 & 69.8 & 47.2 & 48.1 & 48.5 \\
Scene-LLM~\cite{scenellm} & ArXiv24 & 40.9 & \textbf{69.1} & 45.0 & \textbf{70.8} & 47.2 & 52.3 & 54.2 \\
\textbf{Ours} & -- & \textbf{45.38} & 67.02 & \textbf{52.04} & 69.52 & \textbf{49.85} & \textbf{54.95} & \textbf{54.57} \\
\bottomrule
\end{tabular}}
\end{table}

\section{Qualitative Analysis}

In this section, we perform a comprehensive qualitative analysis of ScanQA~\cite{scanqa} and ScanRefer~\cite{scanrefer}. Additional analysis of other datasets will be included in the final version.

\begin{figure*}[tbp]
    \centering
    \includegraphics[width=\textwidth]{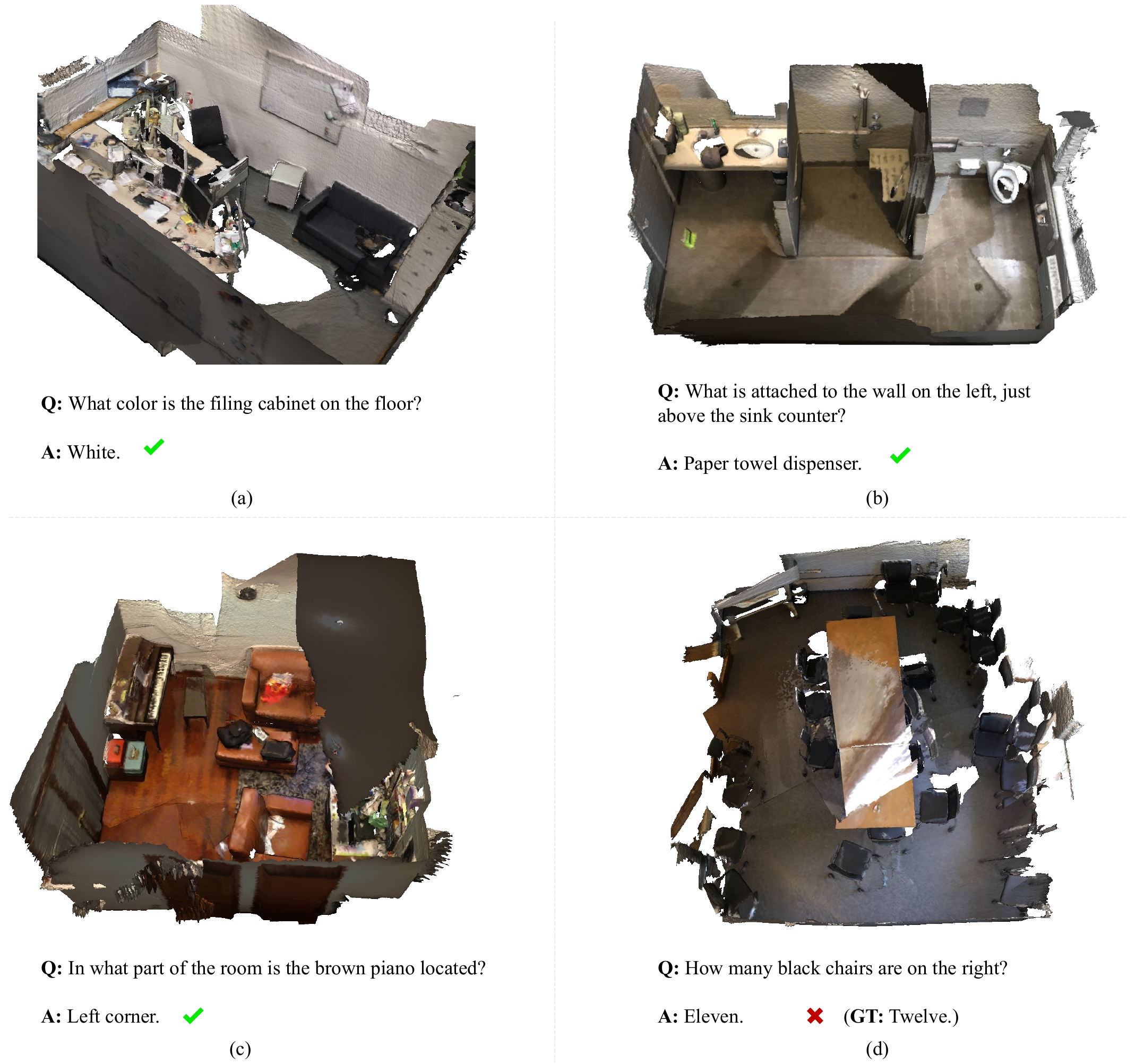}
    \caption{\textbf{Visualization results of 3D question answering on ScanQA~\cite{scanqa}. }}
    \label{fig:scanqa_vis}
\end{figure*}

\subsection{3D Question Answering}
We present four evaluation results for 3D question answering on ScanQA~\cite{scanqa} dataset, as shown in Figure~\ref{fig:scanqa_vis}. In this dataset, both the input and the output do not contain object referencing. This dataset does not include object referencing in either the input or output. Example (a) necessitates the model's ability to perceive an object's appearance, specifically its color. Example (b) demands that the model identify a target object based on a descriptive prompt. Example (c) involves the model describing the position of a target object, while example (d) tests the model's capability to count objects. Our model demonstrates relatively high performance on the first three types of tasks but often struggles with the fourth, particularly when the count of target objects is large. Accurately perceiving and localizing each object is essential for the counting task; failure to do so results in incorrect answers. This challenge persists in both our method and previous methods. Moreover, the inferior annotation quality within the ScanQA dataset exacerbates this issue. For instance, the question in example (d), ``How many black chairs are on the right?'' lacks a precise definition of ``right'', leading to potential confusion for the model.

\begin{figure*}[tbp]
    \centering
    \includegraphics[width=\textwidth]{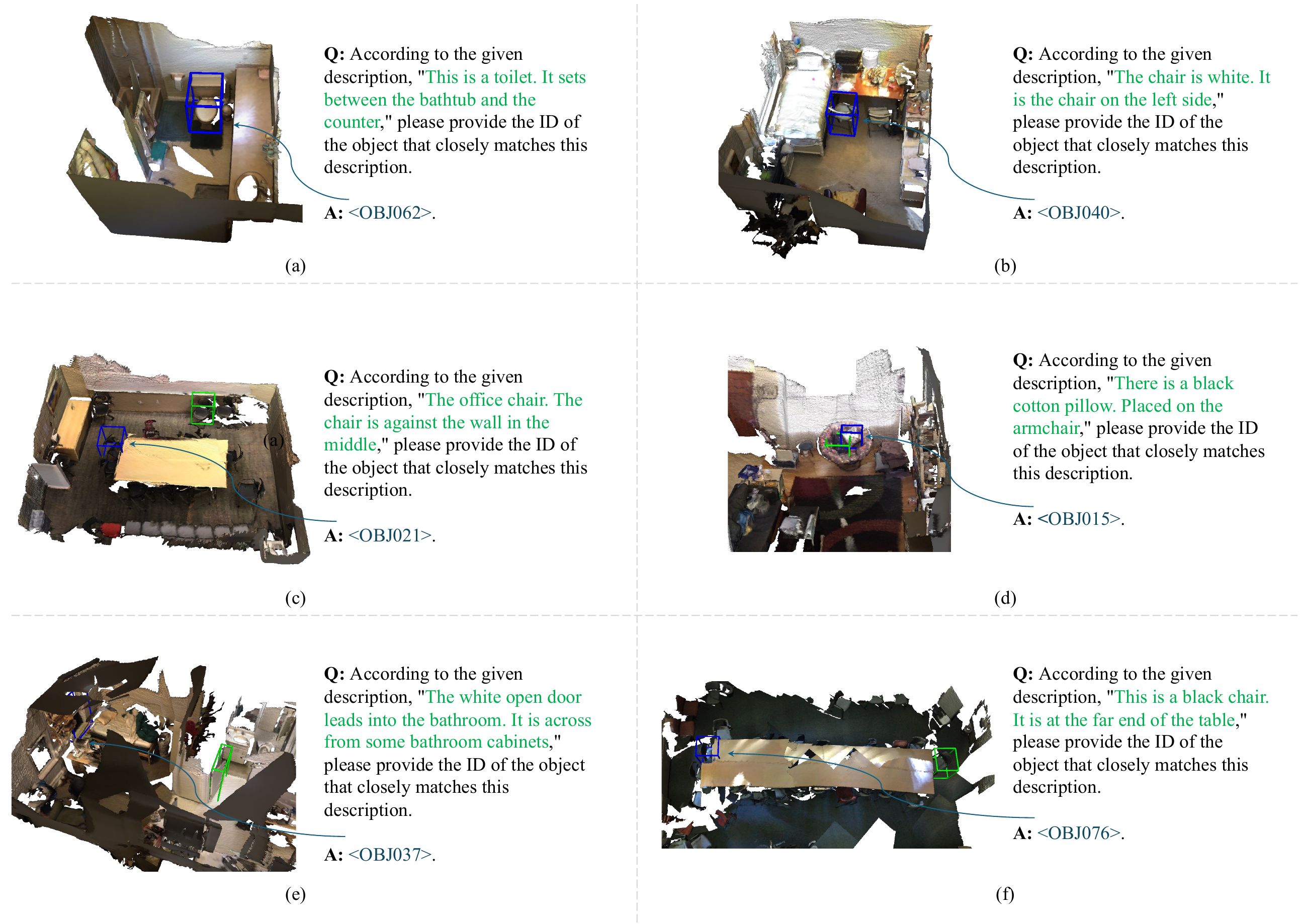}
    \caption{\textbf{Visualization results of 3D visual grounding on ScanRefer~\cite{scanrefer}.} The predicted {\color{blue}blue} box is transformed from the segmented point cloud of the predicted object (ID). The {\color{green} green} ground truth box is provided when the prediction is wrong.}
    \label{fig:scanrefer_vis}
\end{figure*}

\subsection{3D Visual Grounding}
We present six evaluation results for 3D visual grounding on ScanRefer~\cite{scanrefer} dataset, as illustrated in Figure~\ref{fig:scanrefer_vis}. This task challenges the model to localize a target object based on a descriptive prompt. For simpler scenarios, such as those in examples (a) and (b), our model performs adequately. However, it struggles with the remaining four examples for various reasons.

In example (c), the model is tasked with identifying a chair ``against the wall'' but erroneously selects a chair that is not positioned as described. This highlights a deficiency in our model's understanding of interior structural elements like walls, ceilings, and floors. Despite the presence of segmented annotations for these surfaces, they are typically not utilized in training because they are not considered objects per se. This limitation is likely shared by many current methods. Future work is necessary to enhance the model's recognition of these elements, given their significance in comprehending the entirety of a 3D scene. In example (d), the challenge involves identifying two pillows ``placed on the armchair', one black and the other white. The model correctly locates a pillow on the armchair but fails to distinguish it by color. Example (e) presents a scenario where the model confuses a window for a door, likely due to their similar appearances and the often incomplete nature of the input point cloud. In example (f), the model's selection meets the description, illustrating a flaw in the ScanRefer dataset annotations: some descriptions may correspond to multiple objects, rendering them ineffective for evaluating the visual grounding of a single object.

\begin{figure*}[tbp]
    \centering
    \includegraphics[width=\textwidth]{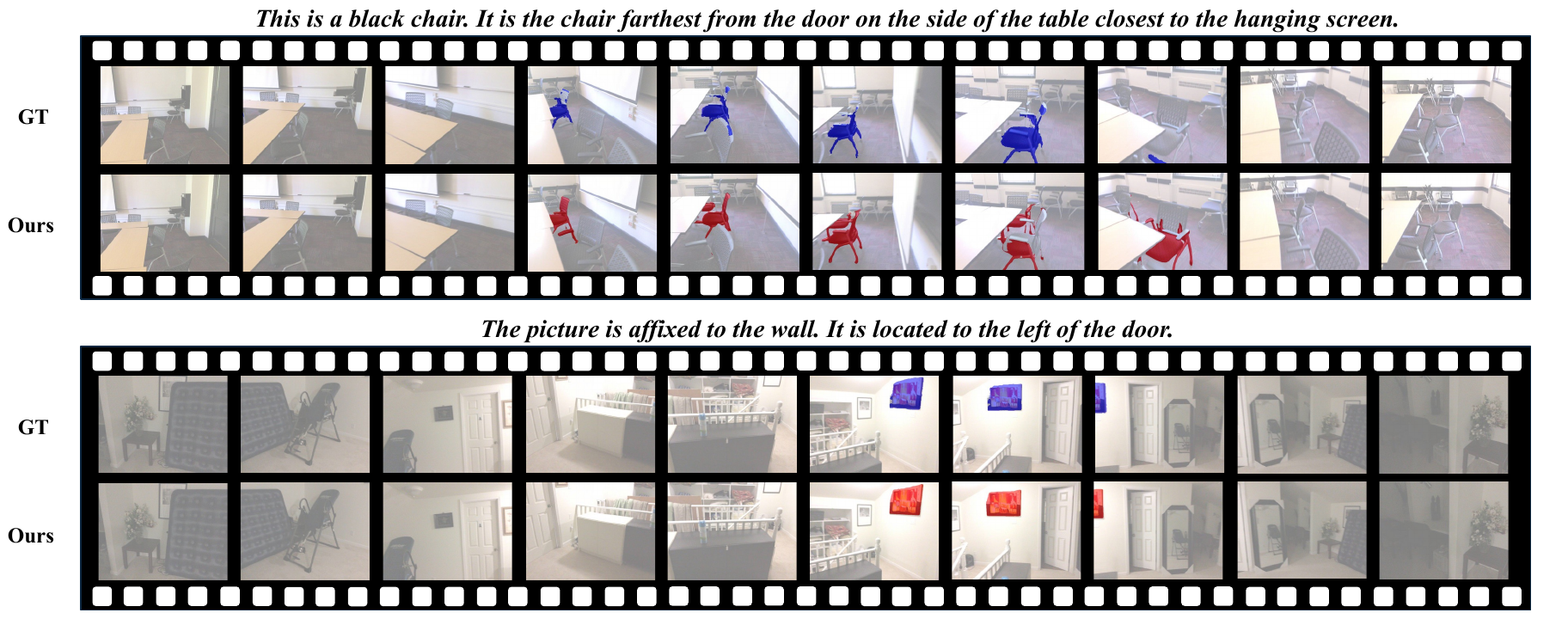}
    \caption{\textbf{Additional visualization results of video grounding for video input.}}
    \label{fig:scanrefer_vis}
\end{figure*}




\section{Input/Output Format Comparison}
We present comparative examples of input/output formats in Figure~\ref{fig:format_comparison}. We juxtapose our model against leading \textbf{expert models} (3D-VisTA~\cite{3dvista}, 3D-VLP~\cite{3d-vlp}, 3DJCG~\cite{3djcg}, M3DRef~\cite{multi3drefer}, and MVT~\cite{mvt}) as well as \textbf{3D MLLMs} (3D-LLM~\cite{3dllm}, Chat-3D~\cite{chat3d}, LLM-Grounder~\cite{llmgrounder}, LL3DA~\cite{ll3da}, and LEO~\cite{leo}). We enumerate several common tasks for varied combinations of input and output formats and provide examples of interactions using object identifiers with our model. It is necessary to acknowledge that some responses shown here are not directly produced by our model due to a lack of adequate data for fine-tuning on these tasks, such as multi-object summaries, multiple-choice QA, and complex planning. The comparison underscores the potential for employing object identifiers to address complex tasks involving single or multiple object references in the input/output, representing a significant enhancement over previous methods restricted to simple tasks with basic formats.

\begin{table*}[tbp]
    \centering
    \caption{\textbf{Comparison of input/output formats.} The use of object identifiers make it possible to solve tasks with {\color{orange}single} or even {\color{red}multiple} object reference in the input/output.}
    \includegraphics[width=\textwidth]{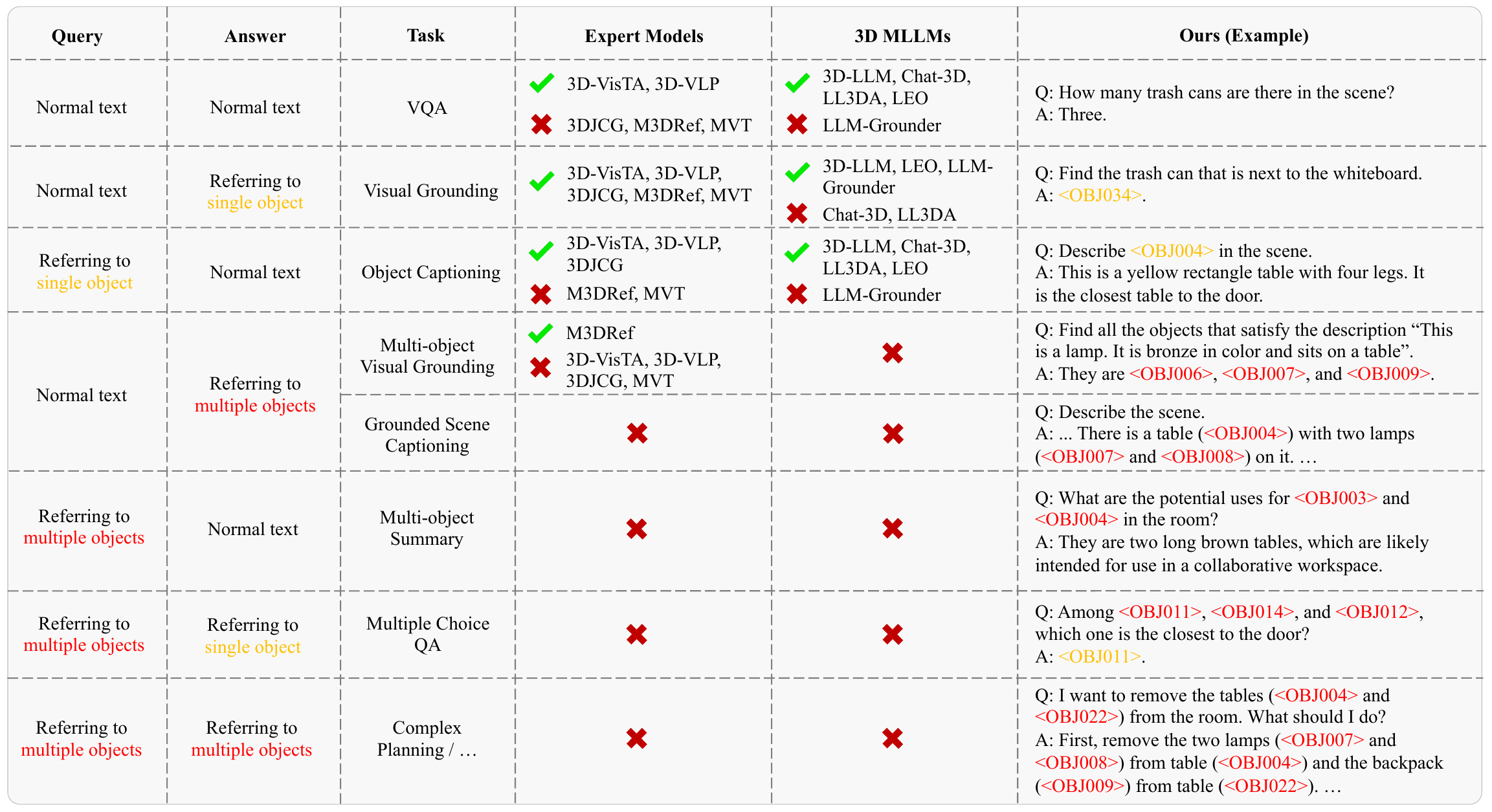}
    \label{fig:format_comparison}
\end{table*}

\begin{table*}[tbp]
\caption{\textbf{Prompt templates for different tasks.} <Description> is replaced by an object's description in ScanRefer/Multi3DRefer/Scan2Cap. <Question> and <Answer> denotes the question and answer in ScanQA/SQA3D. <Situation> is the situation in SQA3D.}
\begin{tcolorbox}[colback=lightgray!10,
                    colframe=black,
                    width=\textwidth,
                    arc=1mm, auto outer arc,
                    boxrule=0.5pt,
                    ]
{\color{NavyBlue}Single-object Visual Grounding (ScanRefer):}

\textbf{User}: What is the ID of the object that matches the description ``<Description>''?

\textbf{Assistant}: {\color{OrangeRed}$\texttt{<OBJXXX>}$}.

\tcbline

{\color{NavyBlue}Multi-object Visual Grounding (ScanRefer):}

\textbf{User}: Are there objects described as ``<Description>''? If there are, please provide the IDs for those objects.

\textbf{Assistant}: Yes. {\color{OrangeRed}$\texttt{<OBJXXX>}$}, {\color{OrangeRed}$\texttt{<OBJXXX>}$}, and {\color{OrangeRed}$\texttt{<OBJXXX>}$}.  {\color{NavyBlue} (for matched cases)}

\textbf{Assistant}: No.  {\color{NavyBlue} (for unmatched cases)}

\tcbline

{\color{NavyBlue}Dense Captioning (Scan2Cap):}

\textbf{User}: Provide a detailed description of the appearance of {\color{OrangeRed}$\texttt{<OBJXXX>}$} before analyzing its spatial connections with other elements in the scene.

\textbf{Assistant}: <Description>

\tcbline

{\color{NavyBlue}Visual Question Answering (ScanQA):}

\textbf{User}: <Question> The answer should be a phrase or a single word.

\textbf{Assistant}: <Answer>

\tcbline

{\color{NavyBlue}Situated Question Answering (SQA3D):}

\textbf{User}: <Situation> <Question> The answer should be a phrase or a single word.

\textbf{Assistant}: <Answer>

\end{tcolorbox}
\label{tab:prompt_example_for_tasks}
\end{table*}

\section{Limitation and Societal Impact\label{appendix:limitation}}
\noindent\textbf{Limitation.}
The primary limitation of our method stems from its reliance on pre-trained foundation models, including 2D/3D detectors and 2D/3D encoders. In our experiments, we froze these models under the presumption of their robust performance. Yet, they occasionally produce incorrect results, as evidenced by the failure cases detailed in the supplementary material. To establish an end-to-end pipeline and enhance model performance further, future work will involve integrating these foundation models into the entire training process.

Another significant challenge is the scarcity of data. While the development of 2D vision-language models has benefited from the availability of millions of image-text pairs for pre-training, the 3D-language domain grapples with a dearth of corresponding data, adversely affecting the alignment between 3D and language spaces. This issue is particularly acute in 3D scene understanding, where the limited availability of scene-language pairs restricts training by failing to provide adequate spatial relationship data. Despite our model's impressive performance in various evaluations, it occasionally misclassifies objects, notably in underrepresented classes such as "hair dryer" and "soap dish." Future initiatives should aim to enhance data volume to bolster the 3D MLLM’s scene understanding capabilities.

\noindent\textbf{Societal Impact.}  Our model has achieved consistent performance improvements in multiple tasks related to 3D indoor scene understanding, which is beneficial and potentially applicable to downstream applications. However, due to the training data not covering all possible scenarios, the model may experience hallucinations during prediction, thereby posing some risks in system applications.

\end{document}